# Active Learning Enables Generation of Molecules that Advance the Known Pareto Front

Evan Antoniuk[1],* Peggy Li[1], Nathan Keilbart[1], Stephen Weitzner[1], Bhavya Kailkhura[1], Anna M. Hiszpanski[1]*

[1]Lawrence Livermore National Laboratory, Livermore, USA
*Corresponding authors: antoniuk1@llnl.gov; hiszpanski2@llnl.gov

## Abstract

Although generative models hold promise for discovering molecules with optimized desired properties, they often fail to suggest synthesizable molecules that improve upon the properties of the structures represented in the training distribution. We find that this limitation arises not only from the molecule generation process itself, but also from the poor generalization capabilities of molecular property predictors. We address this challenge by creating a closed-loop molecule generation pipeline with iterative retraining on new quantum chemical simulation data. Compared against static, single-pass generative modeling approaches, only our closed-loop iterative workflow generates molecules with properties extending beyond the training distribution (up to 0.44 standard deviations beyond the original range) and achieves a 79% improvement in out-of-distribution molecule classification accuracy. Furthermore, by conditioning molecular generation on thermodynamic stability data obtained during the iterative loop, the proportion of stable and hence potentially synthesizable molecules generated is 3.5x higher than the next-best model.

## Introduction

Early efforts in applying machine learning for accelerating new molecule discovery have largely focused on forward-predictive models that output predicted properties of interest given molecules as inputs[1–4]. Molecule discovery can then be conducted by rapidly screening databases or datasets to identify known or proposed molecules with desirable properties[5–8]. However, this screening approach is inherently limited by the size of the screening dataset. Whereas small molecule datasets typically contain $10^6$-$10^9$ entries,[2,9,10] the entire chemical space has been estimated to enumerate up to $10^{60}$ molecules, which is prohibitively large for brute-force screening.[11] More recently, generative or inverse-design models have been proposed as a new paradigm for materials discovery due to their ability to efficiently navigate chemical space beyond what is present in existing databases.[12,13]

The goal of property-constrained molecular generation is to generate novel molecules that possess desirable properties for the application of interest. Typically, a ground-truth oracle function is defined for each molecule design task to quantitatively assess how well the generated molecules meet the desired molecular properties. As a means to quickly approximate this oracle function, property prediction models are used as a surrogate model. These property prediction models are first trained on a pre-existing dataset of molecular properties to learn the mapping between the chemical structure of the molecules and their target molecular properties. After training, this property prediction model is then used to steer the generative model to suggest novel molecules that satisfy the required properties. A wide range of such goal-oriented molecule generative models have emerged within the last 6 years alone, including variational autoencoders (VAEs),[14,15] genetic algorithms,[16–18] reinforcement learning,[19] diffusion models,[20] and chemical language models.[21]

Within the context of material optimization, the Pareto front is the set of materials that possess the optimal properties, such that improving one target property negatively impacts the others. Despite the rapid development of molecular generative models, they have yet to consistently generate state-of-the-art molecules that extend beyond the Pareto front of the training data, to the best of our knowledge. Specifically, across two molecular design benchmarks for organic photovoltaic molecules, none of the eight generative models produced novel molecules that significantly outperformed known molecules in the training set.[22,23] Similarly, prior work has shown that Bayesian optimization-based molecule generation fails to generate valid molecules when the generated molecules are located far from the training molecules in latent space.[24] Although this can be mitigated by constraining the generative model to only sample regions of chemical space that are well represented by the training data,[24,25] constraining the generative model in this way will prevent discovering exciting molecules in new and unexpected regions of chemical space.

We propose that the limited ability of molecular generative models to generate Pareto optimal molecules arises not from the molecule generator itself, but from the property prediction model's failure to generalize reliably to new regions of chemical space. By design, it is the job of the generative model to generate out-of-distribution molecules that have properties that extend beyond what is present in the training data. However, a fundamental principle of regression models, including the property prediction model that guides the molecular generation, is that they will not extrapolate well beyond their training data.[26,27] If the property prediction model that is guiding the molecular generation cannot generalize to out-of-distribution molecules, we propose that the molecular generation model will also fail to generate molecules with properties exceeding that of the training data (Figure 1a). An alternative approach to overcome the limited extrapolative capabilities of regression models is to omit property prediction models entirely and run physical simulations on all generated molecules to directly obtain their properties. Although this approach may be possible for some simple properties, it is prohibitively computationally expensive in general.

Existing molecular generative models also struggle to generate molecules that can be experimentally synthesized.[28] Previous attempts to incorporate synthesizability constraints into molecular generation have explored the incorporation of synthesizability scores (such as SAScore or SCScore)[29,30] or the use of computer-assisted synthesis planning (CASP) tools into the molecule generation process.[31–33] Although CASP tools that predict a complete retrosynthesis pathway are typically too computationally expensive to use within a generative model, recent generative workflows have incorporated synthesizability scores[34,35] or enforce a strict oracle budget of only generating 1000 molecules.[28,36] Generally, all synthesizability scores are hindered by the limited range of known molecules that they are trained on.[29,30,37] This is likely to limit the discovery of new chemical moieties since the generation process will be biased towards domains of already known chemistry.

Several recent works have highlighted the acceleration in materials discovery that can be achieved through the development of closed-loop, active-learning workflows that couple expensive physical simulations with machine learning.[38–40] In the context of material discovery, we here adopt the definition of active learning from Bilodeau et al. as “an experimental strategy that involves iteratively selecting new experiments to perform with the goal of minimizing the cost of acquiring training data while simultaneously maximizing the accuracy of the machine learned model”. Although these prior works highlight active learning’s acceleration in molecular discovery, the improvement in the ability of the entire generative model pipeline to generate molecules that extend the known Pareto front have yet to be explored.

In this work, we show that a simple way to enable molecular generative design models to generate molecules beyond their training distribution is to couple generative modeling with a closed-loop, iterative retraining process using high-throughput quantum-chemical simulations. Within our active learning pipeline, new molecules suggested by the generative model have their properties and stability verified by accurate density functional theory (DFT) simulations. We emphasize that our active learning approach does not rely on uncertainty quantification. Rather, it can be best described as employing diversity-based sampling in the molecular property space to select new molecules for verification with DFT simulations. The results of these ab-initio simulations are then used to retrain the property prediction models, such that properties of molecular candidates in new regions of chemical space are verified and errors in the property prediction models are self-corrected. By focusing on how the generated molecular candidates improve the generalizability of the property prediction surrogate model, we thereby elucidate how active learning enables exploration in regions of chemical space not seen in the original training dataset.
Our work results in three main contributions:

i) We show that including active learning on quantum simulations in closed-loop molecular generation outperforms existing molecule generative tools both in terms of generating molecules with superior properties, as well as generating Pareto-efficient molecules with high consistency. Among all tested generative models, the ability to generate molecules that advance the known Pareto front in a multi-property molecule optimization task was only achieved through the inclusion of active learning.

ii) We show that a key failure mode of existing generative models is that their property prediction models fail to generalize to regions of new chemical space not seen in training. We find that iterative active learning in the new chemical space is a powerful strategy for improving predictive performance, resulting in up to a 19x reduction in the property prediction RMSE on the generated molecules. Of particular interest to generative modeling, we show that retraining the property prediction model improves its precision for identifying top-performing molecules from 7% to 86%.

iii) We also show that conditioning the molecular generative model on the thermodynamic stability data from prior DFT-relaxed generated molecules greatly improves the fraction of stable generated molecules. To accomplish this, we train a molecule graph neural network classifier to filter out unstable generated molecules, which improves the fraction of stable generated molecules to be 3.5x higher than the next best generative model.

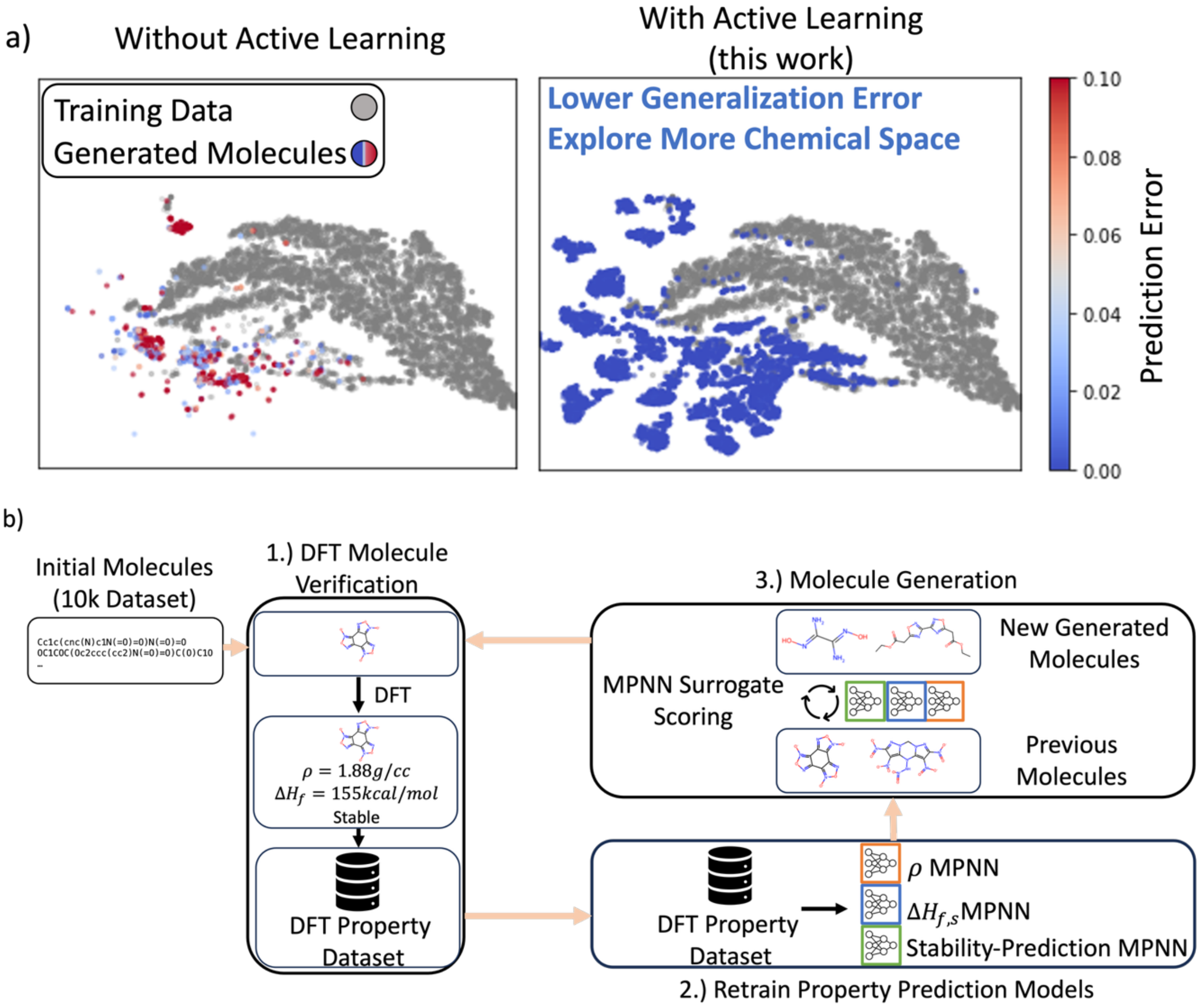

**Figure 1.** a) T-SNE visualization of the molecules generated in this work without active learning (left) and after four iterations of active learning (right). Generated molecules are colored by the absolute error between their DFT calculated density and their density predicted by a Message-Passing Neural Network (MPNN) model. In the left case, the MPNN model is only trained on the 10k Dataset, whereas the MPNN model in the right case is trained after three iterations of active learning, as described in Figure 1b. b) Pipeline for molecule generation with active learning. Starting from an initial dataset that spans the chemical search space of interest (10k Dataset), our pipeline iteratively follows the following cycle of steps. 1.) The molecular properties of interest (density and solid heat of formation) and the thermodynamic stability of all molecules are calculated with DFT. 2.) We train three separate MPNN models on all molecules that have been passed through the DFT calculations. Density (ρ) and solid heat of formation ($\Delta H_{f,s}$) MPNNs are trained as regression models only on values from stable

molecules, whereas the Stability-Prediction MPNN model is trained to classify between DFT-stable and DFT-unstable molecules. 3.) We generate new candidate molecules with the JANUS genetic algorithm, using the retrained MPNNs to evaluate the generated molecules (Equation 2). Finally, these newly generated molecules are fed back into the DFT calculations to complete the active learning loop.

# Results

## Computational Pipeline Overview

To evaluate the importance of active learning for real-world molecule discovery tasks, we focus on the maximization of two molecular properties: density ($\rho$) and solid heat of formation ($\Delta H_{f,s}$) due to their relevance in a wide range of molecular applications.[41–44] Our initial dataset consists of 10,206 known molecules previously collected from the Cambridge Structural Database (CSD), which we hereafter refer to as the '10k Dataset'.[42] This 10k Dataset represents all known molecules in the CSD that only contain carbon, hydrogen, oxygen and nitrogen atoms, and contain at least one nitrogen-oxygen bond.

Existing property-constrained molecular generative modules typically consist of two main components: a molecular generative model and a property prediction model.[13] The molecular generative model outputs molecular structures, whereas the property prediction model evaluates the properties of the generated molecules. The outputs of the property prediction model evaluations are then fed back into the molecular generative model to steer the generative model towards promising molecular candidates. This standard framework has two notable limitations. First, there is no explicit check to ensure that the generated molecules are synthesizable. Second, since the property prediction model is guiding the molecular generation, any misclassifications made by this property prediction model will push molecular generation towards unfruitful regions of chemical space without any method for self-correction.

In this work, we build upon this standard molecular generation workflow by including a third component: a DFT pipeline for validating molecular properties and stability that is included in an active learning fashion (Figure 1). Specifically, after a batch of new molecules is generated, the DFT pipeline determines the relaxed 3D geometry of the molecule, the molecule's $\rho$, $\Delta H_{f,s}$, and its stability. Then, these molecules and their corresponding properties are used to retrain three separate Message-Passing Neural Network (MPNN) models: one for $\rho$, $\Delta H_{f,s}$, and a stability classifier, that we hereafter call the Stability-Prediction MPNN model. This Stability-Prediction MPNN is trained on all previous DFT relaxations, where the thermodynamically stable/unstable molecules are treated as positive/negative examples, respectively. Notably, this active-learning loop ensures that the molecules proposed by the generative model are immediately validated by DFT for thermodynamic stability. Additionally, any misclassifications made by the property prediction models are self-corrected by the DFT-calculated property values, thereby improving the performance of the MPNN models in new chemical space. Overall, our molecular generation workflow is analogous to the approach of Loeffler et al., whereby the binding free energy molecular dynamics simulations in that work are replaced by our DFT pipeline for predicting the molecule's $\rho$ and $\Delta H_{f,s}$.[45]

We use the JANUS genetic algorithm as the molecular generative model due to its recent state-of-the-art performance across multiple inverse design benchmarks.[17] JANUS maintains two fixed-size populations of molecules: an exploration population that broadly searches chemical space and an exploitation population that finetunes within regions of chemical space with high scoring molecules (Methods). All generated molecules are evaluated by a multi-property optimization score (Methods, Equation 2).

## Active Learning Procedure

For all molecules in the 10k Dataset, we calculate both $\rho$ and $\Delta H_{f,s}$ with DFT. Then, we train MPNN models on the DFT-calculated $\rho$ and $\Delta H_{f,s}$ values of the 10k Dataset. Following this initialization, we perform four total iterations of active learning, as summarized in Table 3 (Methods). After each iteration, we retrain the MPNN prediction models on the DFT-calculated $\rho$ and $\Delta H_{f,s}$ values of all molecules generated in all previous iterations, as well as the 10k Dataset. We denote a MPNN model as $MPNN_x$ to refer to the MPNN model that was trained on the 10k Dataset plus the first *X* Iterations of generated molecules. During the first three iterations, we improve molecules' chemical diversity by randomly sampling molecules with both high and low property values to validate with DFT. We show in Figures S18-S19 that this random sampling of generated molecules leads to improved MPNN performance, when compared to sampling based on chemical diversity or top molecule properties. In the fourth and final iteration, we only select the 500 molecules with the highest MPNN-predicted $\rho$ and $\Delta H_{f,s}$ for DFT evaluation. As a result, we describe our active learning sampling approach as diversity-sampling in the property space, rather than an approach based on model uncertainties. The Stability-Prediction MPNN is used in the fourth iteration to guide the molecular generation towards thermodynamically stable generated molecules (see Methods).

## Active-Learning Enables Generation of Molecules that Advance the Known Pareto Front

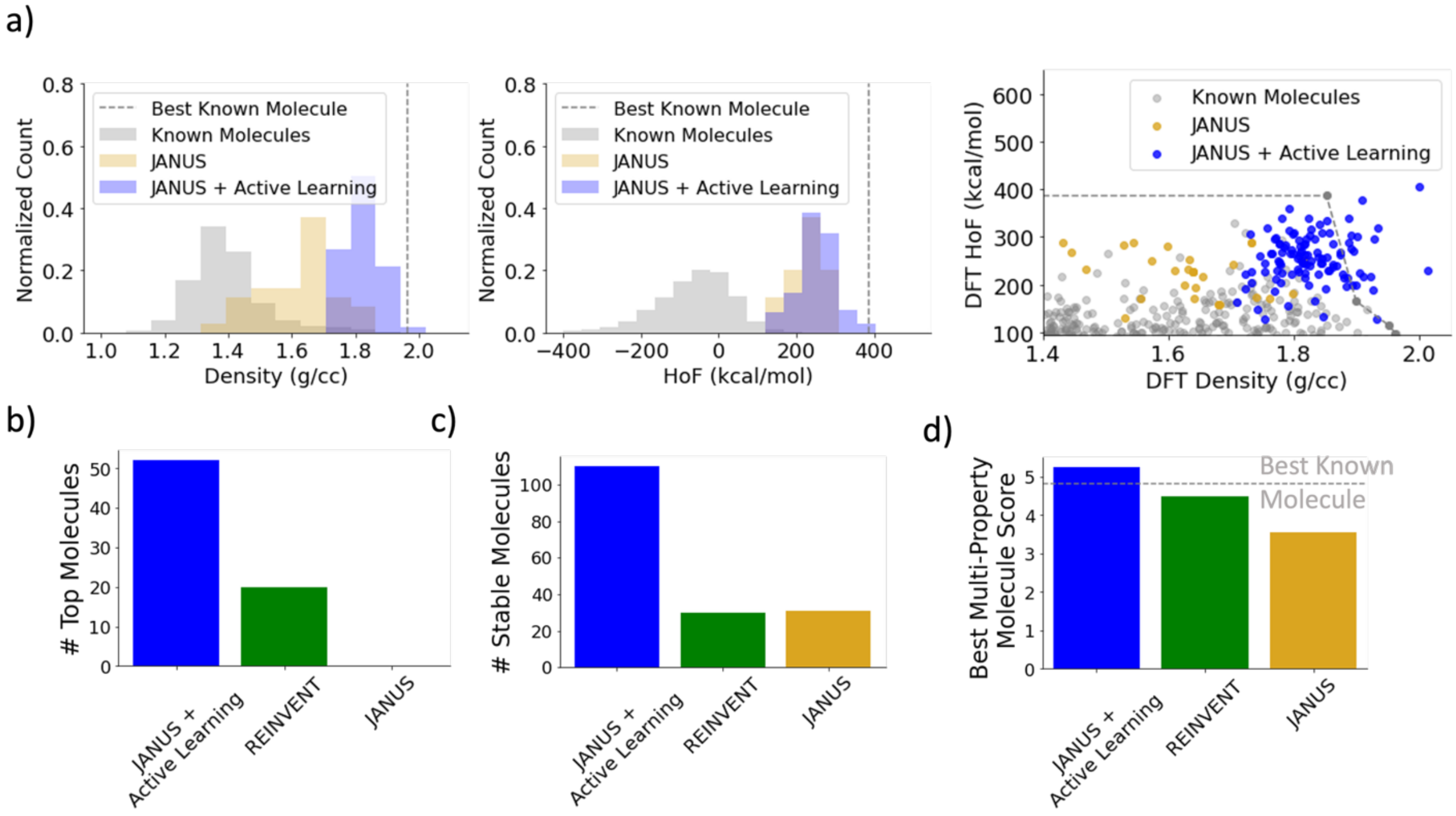

**Figure 2.** Comparison of performance of molecular generation approaches. For all plots, generative models are limited to a DFT calculation limit (oracle budget) of 500 molecules. a) Histograms of generated molecule densities (left), solid heat of formations (middle) and multi-property optimization of both density and heat of formation (right). In the multi-property setting, the Pareto front of the 10k Dataset is shown by the dotted line. All values are calculated with DFT. b) Number of generated molecules that have both high heat of formation and density values, defined as having a value three standard deviations above the mean of the training data. c) Number of stable molecules generated by each generative model. Molecule stability is determined by DFT. d) Single best multi-property optimization score achieved by each model. The multi-property score is defined by Equation 1 in the Methods section.

We benchmark the performance of molecular generation with active learning by comparing its results with screening all molecules in the 10k Dataset, as well as two state-of-the-art molecule generative models (JANUS and REINVENT). Notably, REINVENT was recently the best-performing molecule generation algorithm across 25 different molecule generation methods.[46] The comparison to JANUS without active learning serves as an important ablation study to understand how the inclusion of active learning improves generative model performance.

We evaluate the molecular models according to several evaluation criteria defined in Table 1. Notably, the % state-of-the-art (SOTA) molecules metric allows us to quantify the ability of generative models to consistently generate molecules that advance the Pareto front of molecules seen in training, which is the main draw of molecular generative models. Consistent with recent benchmarking showing that current molecular generative models fail to advance the Pareto front with a limited oracle budget, we limit all models to only generate 500 molecules for DFT evaluation.[46]

**Table 1.** Comparison of molecule design methods for the optimization of molecular density and heat of formation. Best performing approach is bolded.

| Approach | Valid | Top DFT $\rho$ (g/cc) | Top DFT $\Delta H_{f,s}$ (kcal/mol) | Top Multi-Property Score[a] | % DFT Stable Molecules[b] | % Top Molecules[c] | % SOTA Molecules[d] |
|---|---|---|---|---|---|---|---|
| Training Dataset Screening | 1.00 | 1.963 | 387 | 4.80 | 100% | - | - |
| **Generative Models** | | | | | | | |
| JANUS | 1.00 | 1.816 | 290 | 3.56 | 6% (31/500) | 0% (0/500) | 0% (0/500) |
| REINVENT w/ $MPNN_0$ | 1.00 | 1.901 | **417** | 4.48 | 6% (30/500) | 4% (20/500) | 2.40% (12/500) |
| JANUS w/ Active Learning (this work) | **1.00** | **2.014** | 405 | **5.24** | **22% (108/500)** | **10% (52/500)** | **3.40% (17/500)** |

[a] Multi-property score is evaluated according to Equation 1 (see Methods).
[b] We define DFT stable molecules as those that the ground state geometry was successfully optimized, the molecular connectivity did not change during molecular relaxation, and the vibrational analysis of the relaxed molecule structure does not contain any negative vibrational modes.
[c] Top molecules are defined as stable molecules that have both a DFT-calculated $\rho$ and $\Delta H_{f,s}$ that is three standard deviations above the training data
[d] SOTA molecules are defined as stable molecules that exceed the Pareto front of molecules in the 10k Dataset in terms of their $\rho$ and $\Delta H_{f,s}$ values.

Table 1 highlights the importance of active learning for generating molecules with significantly improved properties compared to existing generative models without active learning. Notably, neither of the two benchmark generative models were able to improve upon the best multi-property score (Equation 1) of the training data (4.80), whereas the inclusion of active learning resulted in a top molecule score of 5.24. Similarly, neither of the two benchmark generative models were able to generate molecules with DFT density values larger than the highest density molecule within our 10k dataset (1.963g/cc). However, JANUS with active learning generated molecules with densities exceeding 2g/cc. These results empirically establish the improvement in generating molecules with optimal properties due to the inclusion of active learning in the generative model pipeline. Interestingly, across all metrics in Table 1, a larger performance improvement is achieved by adding active learning to JANUS than by using a more performant generative model (REINVENT). As a result, we find that augmenting molecular generative models with active learning may have a larger impact on constrained molecule generation performance than the development of new molecule generation methodology. As an additional ablation experiment, we also used REINVENT to generate high $\rho$ and $\Delta H_{f,s}$ molecules without using active learning, but using either $MPNN_0$ or $MPNN_3$ as the property prediction model (Table S1). We find that using the $MPNN_3$ model improves the average performance of REINVENT, but not its ability to generate molecules that advance the Pareto front. This finding helps to underscore the main conclusion of our work that it is only through multiple rounds of iterative retraining (i.e. active learning) that the generation of new Pareto optimal molecules is possible.

JANUS with active learning also generates a 3.5x higher proportion of stable molecules than both JANUS and REINVENT. The 3.5x higher rate of stable molecule generation is due to the inclusion of the Stability-Prediction MPNN that learns to identify molecules that were previously determined to be unstable according to DFT (Figure 5). Altogether, we find that active-learning improves the sample efficiency by generating a higher proportion of both stable and top-performing molecules.

## Property Prediction Models Fail to Extrapolate in Chemical and Property Space

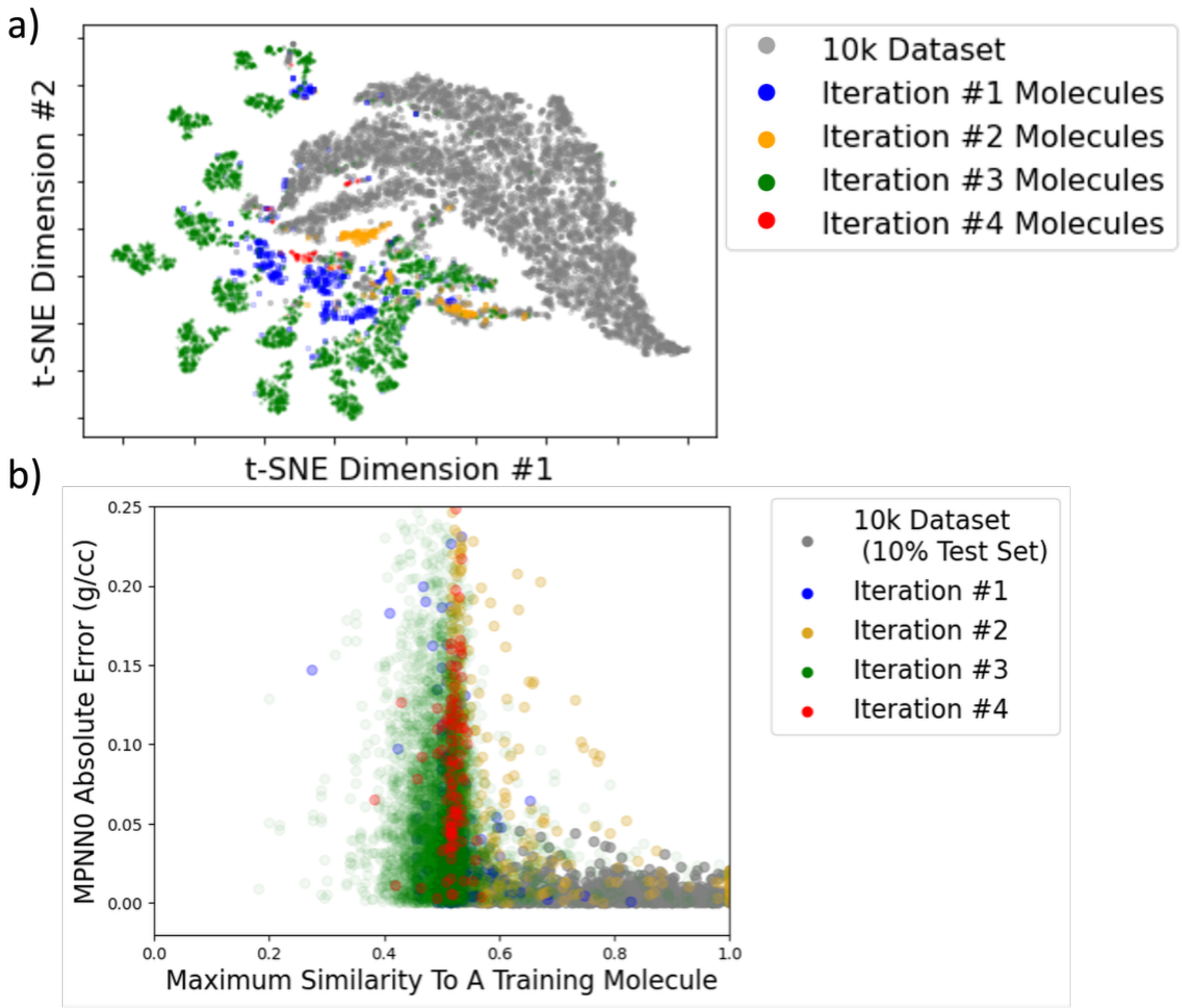

**Figure 3.** Visualization of the molecules generated in the active learning process. a) t-distributed stochastic neighbor embedding (t-SNE) visualization of the 10k Dataset molecules (gray) and each iteration of active learning (colored). RDKit descriptors, as implemented in the DeepChem package,[47] are used to featurize the molecules. b) The error between the $MPNN_0$ predicted

density of a test-set molecule and the DFT-calculated density of the molecule is plotted against the molecule's similarity to molecules in the training dataset. Molecules' similarity to the training dataset is quantified as the Tanimoto similarity between the RDKit fingerprints of each molecule and its nearest molecule in a 90% train split of the 10k Dataset. c) $MPNN_0$ test RMSE for prediction ρ (left) and $\Delta H_{f,s}$ (right) of molecules in the 10k Dataset (in-distribution) and the generated molecules from the four active learning iterations. Parity plots are provided in Figures S2-S11. In all cases, the $MPNN_0$ model is evaluated on a 10% hold-out test set.

The results in Table 1 highlight our key finding that generative models without active learning struggle to consistently generate Pareto optimal molecules. In Figure 3a-b, we visualize how the molecules in the 10k Dataset (known molecules) differ from the generated molecules in the active learning process. Figure 3a qualitatively shows that the generated molecules reside in a significantly different region of chemical space than the 10k Dataset (also see Table S2, Figure S13). We quantify this result in Figure 3b by showing that the minimum distance in chemical space between generated molecules and known molecules in the 10k Dataset is significantly larger than the nearest distance between molecules within the 10k Dataset. Furthermore, Figure 3b shows that the error between the $MPNN_0$ predicted density of the molecule and the DFT-calculated density is only small (<0.1 g/cc) when there are similar enough molecules in the training set (specifically, when the Euclidean distance is less than 30 (arbitrary units)). This result elucidates why including active learning in generative modeling loops is necessary for generating molecules that advance the Pareto front. By continuously retraining the MPNN on molecules from new regions of chemical space, the performance of the MPNN model improves by ensuring that sufficiently similar molecules are present in the training data.

In Figure 3c, we show that the $MPNN_0$ model (without active learning) performs well at predicting $\rho$ and $\Delta H_{f,s}$ for the known test molecules in the 10k Dataset (test RMSE=0.008g/cc and 17kcal/mol, respectively), but fails at predicting the DFT-calculated $\rho$ and $\Delta H_{f,s}$ values of the generated molecules. The resulting prediction error is between 3-11x larger for $\Delta H_{f,s}$ and 11-19x larger for $\rho$ (Figure 3c) compared to the performance on the 10k Dataset test set molecules (Figure 3). We also find that the $MPNN_0$ model exhibits significantly worse predictive performance for property values that differ significantly from the numerical values seen in training (Figure 4b,c). Taken together, these results show that the $MPNN_0$ model struggles to make robust predictions when extrapolating either in chemical space or property space. It is important to note that this poor extrapolation performance is not limited to just the MPNN model. In Figure S12, we also explore the predictive performance of other property prediction models including deep ensemble MPNNs, utilizing larger training sets and state-of-the-art chemical foundation models. All models, none of which use active learning, show poor extrapolation performance.

## Active Learning Improves Property Prediction Models in New Chemical Space

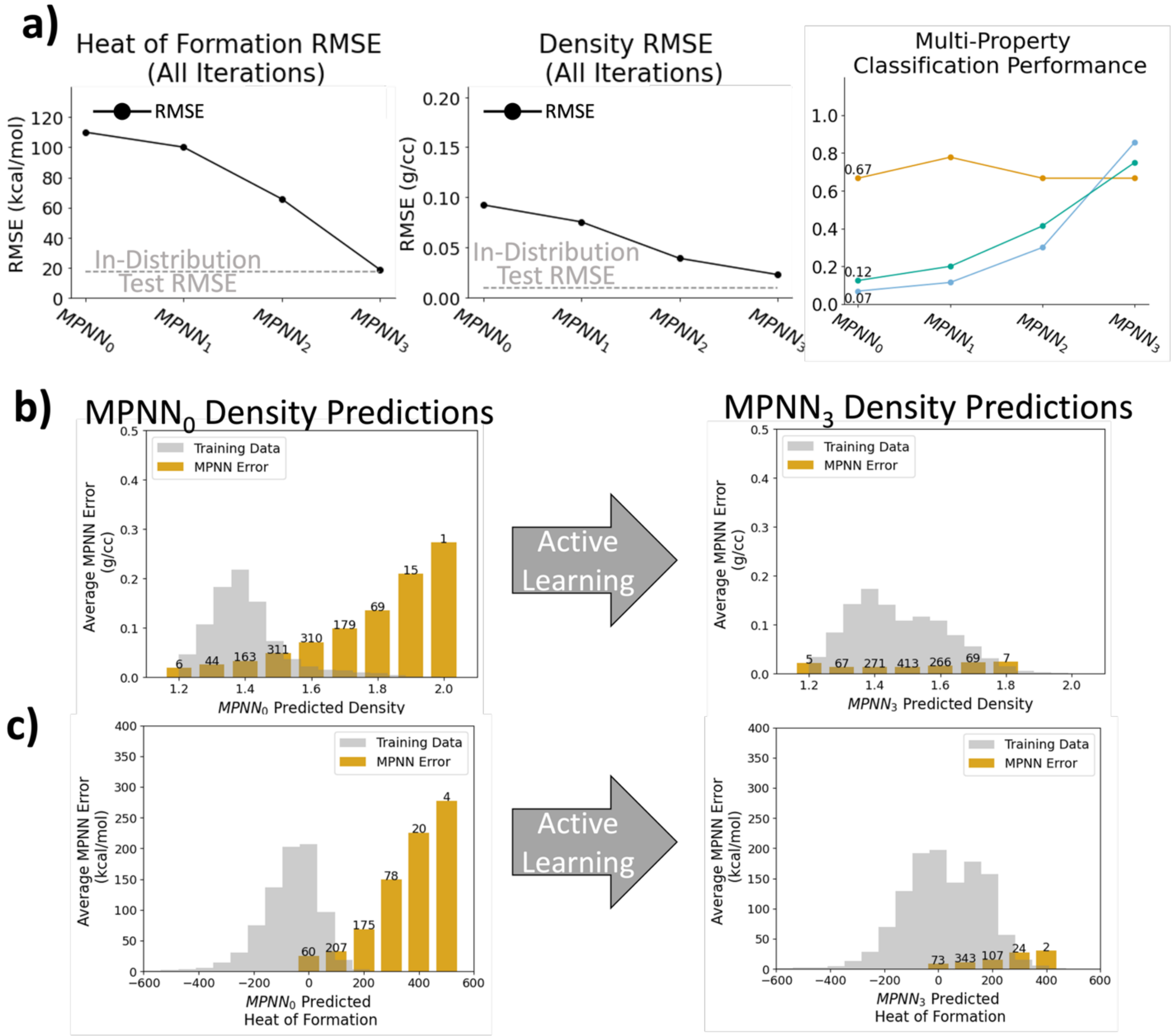

**Figure 4.** a) Regression performance of MPNN models for predicting the density (left) and heat of formation (middle) of generated molecules as a function of active learning iteration. All models are evaluated against a hold-out test set consisting of 10% of the generated molecules from each active learning iteration. On the right, we plot classification performance of these MPNN models for identifying molecules with both high density and heat of formation (defined as three standard deviations above the 10k Dataset mean). b) The orange bars indicate the average absolute error for the MPNN model for molecules with predicted density values within subsequent 0.10g/cc density bins. For example, the first bar indicates that the average absolute error for molecules with predicted densities between 1.2-1.3g/cc is 0.019g/cc. The gray bars indicate the distribution of density values seen in the 10k Dataset (training distribution), normalized to give a maximum bar height equal to half the plot height. The number above each orange bar indicates the number of molecules included in each bin. c) Same as b) but for heat of formation predictions and with a bin size of 100kcal/mol.

In Figure 4a, we show how the property prediction performance of the MPNN model dramatically improves with subsequent iterations of active learning. After three iterations of active learning, the $\Delta H_{f,s}$ prediction RMSE reduces by 83% (from 110kcal/mol to 19kcal/mol) and the $\rho$ prediction RMSE reduces by 75% (from 0.092g/cc to 0.023g/cc), when evaluated on hold-out test molecules from across the entire active learning run. Detailed parity plots are provided in the Supporting Information (Figures S2-11). Notably, for both $\rho$ and $\Delta H_{f,s}$ predictions, the test performance achieved by the $MPNN_3$ models on the generated molecules is comparable to the test performance on the 10k Dataset– indicating successful generalization to the new chemical space (Figure 4a). With active learning, the $MPNN_3$ model achieves a $\rho$ prediction error that is 5x lower than the next-best model, MoLFormer, and a $\Delta H_{f,s}$ prediction error that is 3x lower than MoLFormer (Figure S12).

Understanding how well the MPNN classifies molecules with high $\rho$ and/or $\Delta H_{f,s}$ also elucidates generative model performance since molecular generation is based on the principle that high-scoring molecules will be propagated for further exploration and refinement. The $MPNN_0$ model (without active learning) performs extremely poorly at identifying both generated molecules with high $\rho$ (23% precision) and $\Delta H_{f,s}$ (14% precision) (Figures S6 and S11). This problem is exacerbated in the multi-property setting where the $MPNN_0$ model identifies molecules that have both high $\Delta H_{f,s}$ and $\rho$ with a precision of only 7% (Figure 4a). We show that retraining MPNNs through active learning directly addresses this problem- improving their precision for identifying top performing molecules from 23% to 77% for $\rho$ (Figure S6), from 14% to 81% for $\Delta H_{f,s}$ (Figure S11), and from 7% to 86% for the multi-property setting (Figure 4a). Importantly, we find that active learning greatly improves the precision for identifying top-performing molecules, whereas the recall remains consistently high (Figure 4a). Interestingly, even without active learning, all models do well to correctly recall top molecules. On the other hand, the models trained without active learning have not been exposed to a diverse enough range of high $\rho$ and/or $\Delta H_{f,s}$ molecules, leading to a high rate of false positive predictions. As the models are iteratively retrained, they gain a more precise decision boundary for understanding what specific chemical structures lead to high $\Delta H_{f,s}$ and $\rho$ molecules, thereby improving model precision.

We also visualize how the MPNN prediction errors correlate with the numerical property values (Figure 4b,c). Whereas the prediction error of the $MPNN_0$ model (without active learning) dramatically increases for molecules with larger values of both $\rho$ and $\Delta H_{f,s}$, the $MPNN_3$ model trained with active learning shows low prediction error across all property values. As seen in Figure 4b-c, the training data distribution after three iterations of active learning has provided sufficiently more examples of high $\rho$ and $\Delta H_{f,s}$ molecules, resulting in strong generalization across molecules of any property value.

## Active Learning Improves the Stability of Generated Molecules

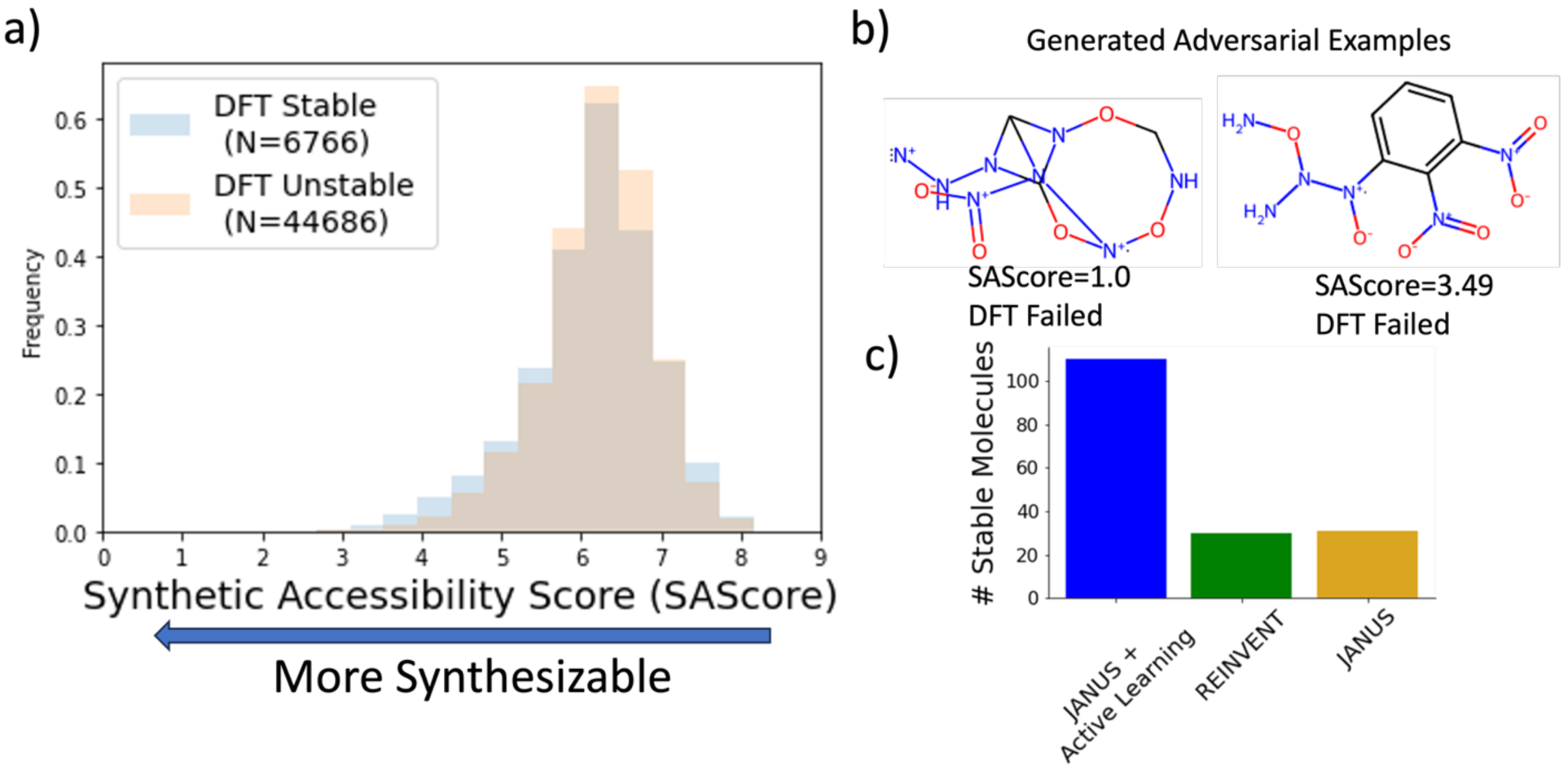

**Figure 5.** a) Distribution of Synthetic Accessibility Scores (SAScores) for all molecules generated throughout the active learning process. b) Examples of adversarial molecules- molecules with high synthetic accessibility scores that are not stable molecules according to DFT. c) Comparison of the number of thermodynamically stable generated molecules from three different generative approaches. All models are constrained to generate exactly 500 molecules.

One of the most important criteria for molecule generation is that the generated molecules must be synthesizable. Although DFT calculations cannot definitively predict if a material is synthesizable, synthesized molecules can be expected to be thermodynamically stable by DFT. Among the 10K Dataset of known, synthesized molecules, 99.6% were found to be DFT stable, indicating that DFT stability is a necessary condition for molecule discovery. As shown in Figure 5a-b, the use of the SAScore has limited utility in discriminating between thermodynamically stable and unstable generated molecules. Using the recommended SAScore cutoff of SAScore<6 to identify synthesizable molecules would result in only 14% of the molecules being thermodynamically stable, which is only marginally better than the random guessing baseline of 13%.

Within our active learning pipeline, we address these shortcomings by training the Stability-Prediction MPNN to steer the molecular generation process towards thermodynamically stable and novel molecules. The Stability-Prediction MPNN achieves a classification AUC of 0.971 at identifying generated molecules that will be unstable according to DFT. In Figure 4c, we compare the fraction of stable generated molecules with REINVENT and JANUS. When active learning is included we generated 110 stable molecules- a 3.5x improvement in the rate of stable molecule generation. As an additional ablation experiment, we also generate 500 molecules with retrained MPNN predictors of $\rho$ and $\Delta H_{f,s}$ ($MPNN_3$), but without the use of the Stability-Prediction MPNN. Under this setting, only 0.4% of the generated molecules were stable, further highlighting the importance of including thermodynamic stability constraints in molecule generation.

We have also performed an ablation experiment where we explore how including various synthesizability scores in the scoring function of the generative model impacts the synthesizability of the generated molecules. Specifically, we explore three experimental settings, i) a baseline where we do not include any synthesizability constraints in the scoring function, ii) including SAScore and iii) using our DFT Stability MPNN (our approach).

**Table 2.** Comparison of generative model performance when various synthesizability constraints are included in the scoring function. Best performing generative approach is bolded for each metric. Additional experimental details are provided in the Methods section. Distributions of the SAScore and RAScores of the generated molecules are provided in Figure S14-15. Additional chemical diversity metrics are provided in Table S3.

| Synthesizability Metric Used in Scoring Function | Top $MPNN_3$ Density (g/cc) | Top $MPNN_3$ $\Delta H_{f,s}$ (kcal/mol) | Top $MPNN_3$ Multi-Property Score | % Synthesizable (SAScore) | % Synthesizable (RAScore) | % DFT Stable (Stability MPNN) |
|---|---|---|---|---|---|---|
| None | **2.06** | **644** | **6.58** | 2/500 | 236/500 | 0/500 |
| SAScore (Equation 3) | 1.98 | 469 | 4.98 | **321/500** | 196/500 | 103/500 |
| DFT Stability MPNN (this work) | 2.03 | 525 | 5.47 | 128/500 | **277/500** | **500/500** |

As shown in Table 2, we find that there is a trade-off between synthesizability and how far in property space the generative models can extend. When no synthesizability constraint is added into the scoring function, we find that JANUS is able to generate molecules with much higher density and $\Delta H_{f,s}$ values, as predicted by the $MPNN_3$ model. However, without enforcing any synthesizability constraints, very few of the generated molecules are predicted to be synthesizable. Table 2 also highlights how our inclusion of the DFT Stability MPNN in the scoring function improves both the properties of the generated molecules and their predicted synthesizabilities, compared to including SAScore in the scoring function. Unsurprisingly, the generative models produce molecules that maximize the synthesizability metric that was directly used in the scoring function. To evaluate molecule synthesizability against a metric not used explicitly in the scoring function, we also calculate the RAScore of the generated molecules. RAScore is a machine learning classifier model that has been trained on the outcomes from the AiZynthFinder CASP tool.[37,48] Notably, even though none of the models directly included RAScore in the scoring function, our approach of using the DFT Stability MPNN produces the highest fraction of synthesizable molecules according to RAScore. Although no approach achieves perfect synthesizability, Table 2 highlights the main contributions of our work that our active learning approach generates molecules with properties that extend beyond the training distribution, while also improving the synthesizability of the generated molecules over baseline generative models.

## Discussion

In this work, we showed that including active learning in molecule generation pipelines vastly improves the performance of the generative model to generate Pareto optimal, stable molecules (Figure 2). Although there has recently been a massive surge in the development of new methodology for molecular generation, our experiments suggest that improving the generalization performance of property prediction models may be even more important for generating novel molecules with state-of-the-art properties. We propose as a best-practice for the field of molecular generative modeling that *all* generated molecules should necessarily be validated through physics-based simulations (such as density functional theory). Since property prediction models do not extrapolate well, reported molecular properties based only on property prediction model predictions alone are likely to be greatly overestimated.

We note that we did not leverage any advanced sampling techniques for determining which molecules will be selected for DFT validation, underscoring the importance of active learning. Nevertheless, we anticipate that sampling molecules based on Bayesian optimization could greatly reduce the number of DFT calculations required.[49] Similarly, further refinement in the number of active learning iterations and number of molecules generated per active learning iteration is likely to reduce the number of DFT calculations required to get comparable performance. Finally, although some molecular properties cannot be simulated rapidly, we anticipate that even a relatively small number of collected molecular property data points could improve the property prediction model generalization performance.

## Methods

### Message-Passing Neural Network

All message-passing neural network (MPNN) models were implemented in the Chemprop package Version 1.4.1.[1,50] This code is available at https://github.com/chemprop/chemprop. Unless otherwise specified, we train all MPNN models on a 80/10/10 train/validation/test split with 5-fold cross validation. For all experiments, we use all default model hyperparameters. All models are trained for 50 epochs and the final model weights for each fold are taken from the epoch that achieved the lowest RMSE on the validation set.

After each iteration of active learning, the MPNNs are re-trained on the molecules that were passed through the DFT calculations. For predicting $\rho$ and $\Delta H_{f,s}$, we only retrain the MPNNs on the 10k Dataset and all generated molecules determined to be thermodynamically stable. The Stability-Prediction MPNN model is trained to classify between the stable/unstable molecules from the first three iterations of active learning.

### JANUS Genetic Algorithm

Within the exploration population, new molecules are obtained by performing mutations (using the STONED algorithm[51] to perform character deletions, additions, and replacements of the molecules' SELFIES string[52]) and crossovers (forming a path between the SELFIES string of two parent molecules in the population and selecting the child molecule along the path that maximizes the joint similarity with both parents). Then, only the molecules with the highest score on the scoring function are propagated to the next generation. Within the exploitation population, new molecules are obtained only by performing mutations. The molecules to be propagated to the next generation are selected based on having high similarity to the parent molecules. Finally, the two populations exchange several high-scoring molecules to facilitate both exploitation and exploration of regions of chemical space with promising candidates.

Generated molecules are evaluated by a multi-property optimization score:

$$Multi - Property\ Score = \left(z_{\Delta H_{f,s}}\right) + \left(z_{\rho}\right) \quad (1)$$

Where $z_{\Delta H_{f,s}}$ is the standard score (z-score) of the molecule's solid heat of formation, and $z_{\rho}$ is the standard score (z-score) of the molecule's density. The mean and standard deviation in the standard score are calculated from the 10k Dataset of known molecules and their DFT-calculated $\Delta H_{f,s}$ and $\rho$. Thus, the multi-property optimization score can be intuitively interpreted as the number of standard deviations by which the target molecule's predicted properties exceeds that of the average molecule in the training data, aggregated across all properties. For example, a molecule with a $\rho$ value 1.2 standard deviations above the training data and $\Delta H_{f,s}$ value 1.5 standard deviations above the training data would have a score of 2.7.

The full objective score that is directly used to guide the generation of JANUS is then given by:

$$Full\ Objective\ Score = X[\sigma(z_{HoF}) + \sigma\left(z_{Density}\right)] \quad (2)$$

Where $\sigma$ is the sigmoid function. In practice, this sigmoid function limits the contribution of each property value to have a maximum value of 1, which is necessary to prevent the molecular generation from being dominated by exceedingly large z-scores arising from erroneously large MPNN predicted property values. Finally, *X* acts to enforce chemical structure constraints by taking on a value of 1 if the molecule meets both the constraints: the molecule only contains C, H, O, and N atoms, and the molecule has a net-zero oxidation state. If both these criteria are not met, *X* has a value of 0. In only the $4^{th}$ iteration of active learning, we expand *X* to include the third criteria that the molecule must be predicted to be stable by the Stability-Prediction MPNN.

As a comparison of our model (using the DFT Stability MPNN as in Equation 2) against using SAScore in the scoring function, we perform an additional round of molecular generation with JANUS with the objective score (Table 2):

$$\text{Objective Score} = [\sigma(z_{\text{HoF}}) + \sigma\left(z_{\text{Density}}\right) + \sigma(z_{\text{SAScore}})] \quad (3)$$

Where we normalize the nominal SAScore range of 1(synthesizable) to 10 (not-synthesizable) to instead range from 0 (not-synthesizable) to 1(synthesizable). In Table 2, We evaluate molecule synthesizability by three metrics to get a more complete picture of synthesizability. For each generated molecule, we classify their synthesizability by SAScore (synthesizable if SAScore<6),

RAScore (synthesizable if RAScore>0.5) and our DFT Stability MPNN (synthesizable if output>0.5).

The JANUS genetic algorithm is adopted from the code is available at https://github.com/aspuru-guzik-group/JANUS.[17] We run JANUS for 200 generations, a generation size of 500, and exchange 5 molecules between the exploitation and exploration populations. Molecules are scored according to the Full Objective Score, detailed below in Equation 2. After running for 200 generations, all generated molecules are collected and filtered to remove any duplicates, molecules with a non-zero formal charge, or molecules that contain atoms other than C,H,N, and O. For active-learning iterations #1-3, we sample molecules from this filtered list for DFT validation, resulting in 980, 2,433, and 48,039 sampled molecules in these first 3 iterations, respectively. The number of sampled molecules in each iteration was chosen based on the availability of computational resources for DFT. For both iteration #4 and JANUS (without active-learning), all generated molecules are collected and filtered as before (to remove any duplicates, molecules with a non-zero formal charge, or molecules that contain atoms other than C,H,N, and O). From this list of filtered molecules, the top 500 molecules to be used for DFT evaluation are determined according to Equation 2. For JANUS with active-learning only, the MPNNs used in calculating the Full Objective Score are re-trained on the DFT-calculated $\Delta H_{f,s}$ and $\rho$ values from all previous iterations and the 10k Dataset. These re-trained MPNNs are trained with 5-fold cross validation and ensembled across all 5 folds to predict the $\Delta H_{f,s}$ and $\rho$ values.

**Table 3.** Overview of Active Learning Molecule Generation Process

| | # Generated Molecules | # DFT Stable Molecules | Property Prediction Model | Uses Stability-Prediction MPNN? | DFT Molecule Selection Process |
|---|---|---|---|---|---|
| Iteration #1 | 980 | 335 | $MPNN_0$ | No | Random |
| Iteration #2 | 2,433 | 362 | $MPNN_1$ | No | Random |
| Iteration #3 | 48,039 | 5,497 | $MPNN_2$ | No | Random |
| Iteration #4 | 500 | 109 | $MPNN_3$ | Yes | Top 500 Molecules |

## REINVENT

We perform all REINVENT experiments using the REINVENT v1.0.1 implementation provided in the Tartarus package.[22,53] The code for this implementation is available at https://github.com/aspuru-guzik-group/Tartarus. REINVENT is initially pretrained using only the SMILES strings from the 10k dataset, and its vocabulary is derived solely from these strings. The scoring function used is the same as JANUS (Equation 2), where the molecular $\Delta H_{f,s}$ and $\rho$ values are obtained from the $MPNN_0$ models, trained on the 10k Dataset. The SMILES vocabulary provided to REINVENT is also derived only from the 10k Dataset. The Stability-Prediction MPNN is not used in molecular generation. The recurrent neural network pretraining was performed for up to 100 epochs with early stopping on an 80% train split of the 10k Dataset. The reinforcement learning agent was then trained with all default hyperparameters (3000 steps with a learning rate of 0.0005 and batch size of 64).

## High-Throughput Density Functional Theory Calculations

We evaluate molecular properties with a high-throughput DFT pipeline (capable of processing thousands of molecules per day) developed within the AiiDA framework.[54,55] Our high-throughput DFT pipeline (capable of processing thousands of molecules per day) was performed with NWChem v7.0.2 and automated using the AiiDA framework for high-throughput simulations.[54–56] Molecular conformations are first generated with RDKit and then optimized using the RDKit force fields. The lowest energy conformation is then used as a starting input for NWChem. Initially, the molecular geometry is relaxed with the B3LYP functional and 6-31G** basis set using tight convergence tolerances. This is then followed by an additional refined relaxation step with the 6-311++G(2d,2p) basis set. Molecules that could not be successfully relaxed into a stable molecular structure are considered to be unstable. Additionally, a connectivity matrix is created for the bonded atoms. If at any point during the structural optimization bonds are broken or created the molecule is considered to deviate from the originally provided SMILES string and discarded from the dataset. For all remaining stable molecules, we then calculate the vibrational frequencies to ensure stable molecules. Molecules containing imaginary frequencies are then removed from the dataset. Finally, we use the methodology of Byrd and Rice for converting quantum mechanical molecular energies of gas molecules to condensed phase heats of formation.[41] The agreement between these DFT-calculated densities and experimentally measured densities are illustrated in Figure S1.

As outlined by Byrd and Rice, to compute the heat of formation of a solid we apply Hess's law which states

$$\Delta H^{o}_{f(s)} = \Delta H^{o}_{f(g)} - \Delta H_{sub}$$

where $\Delta H^{o}_{f(s)}$ is the heat of formation for a solid, $\Delta H^{o}_{f(g)}$ is the heat of formation for a gas, and $\Delta H_{sub}$ is the heat of sublimation. The heat of sublimation is

$$\Delta H_{sub} = a(\mathrm{SA}) + b\sqrt{\sigma^2_{tot}\nu} + c$$

where SA is the surface area at $10^{-3}$ electron/bohr$^3$ isosurface of the electron density, $\sigma^2_{tot}$ is the variability of the electrostatic potential at the same isosurface, and $\nu$ is the balance between the positive and negative charges of the isosurface. The values of *a*, *b*, and *c* are calculated using a least-squares fit of $\Delta H_{sub}$ using experimental values. The equations for computing $\sigma^2_{tot}$ and $\nu$ are provided by Politzer *et al.*.[57] These values are computed using cube files of the electron density and electrostatic potential. A value of $10^{-3}$ electron/bohr$^3$ is used for the isosurface on the electron density. These points are then mapped onto the electrostatic potential and used within the formulation of Byrd and Rice and Politzer *et al.*.[41,57] For the purposes of training the Stability-Prediction MPNN, any molecules for which a stable geometry could not be found or with imaginary frequencies are labelled as unstable molecules.

## Data Availability

The molecules in the 10k Dataset and molecules generated in the first three iterations of active learning are provided in the Supporting Information, along with their DFT calculated solid heat of formation and density values. The 10k Dataset is also available at https://github.com/FLASK-LLNL/LLNL-10k-Dataset.

## Code Availability

The code for the JANUS and REINVENT generative models are available at https://github.com/aspuru-guzik-group/JANUS and https://github.com/aspuru-guzik-group/Tartarus, respectively. The code for the Chemprop MPNN is available at https://github.com/chemprop/chemprop.

## Acknowledgements

This work performed under the auspices of the U.S. Department of Energy by Lawrence Livermore National Laboratory under Contract DE-AC52-07NA27344, document release number LLNL-JRNL-2001596.

## Author Contributions

E.R.A wrote the manuscript and prepared all figures. All machine learning models were developed by E.R.A. DFT calculations were performed by N.K. A.M.H. conceived of the project. E.R.A., P.L., N.K., S.W., B.K. and A.M.H all contributed to reading and editing the manuscript and have approved the manuscript.

## Competing Interests

The authors declare no competing interests.

**Supplementary Information for "Active Learning Enables Generation of Molecules that Advance the Known Pareto Front "**

Evan Antoniuk,* Peggy Li, Nathan Keilbart, Stephen Weitzner, Bhavya Kailkhura, Anna M. Hiszpanski*

Corresponding authors: antoniuk1@llnl.gov; hiszpanski2@llnl.gov

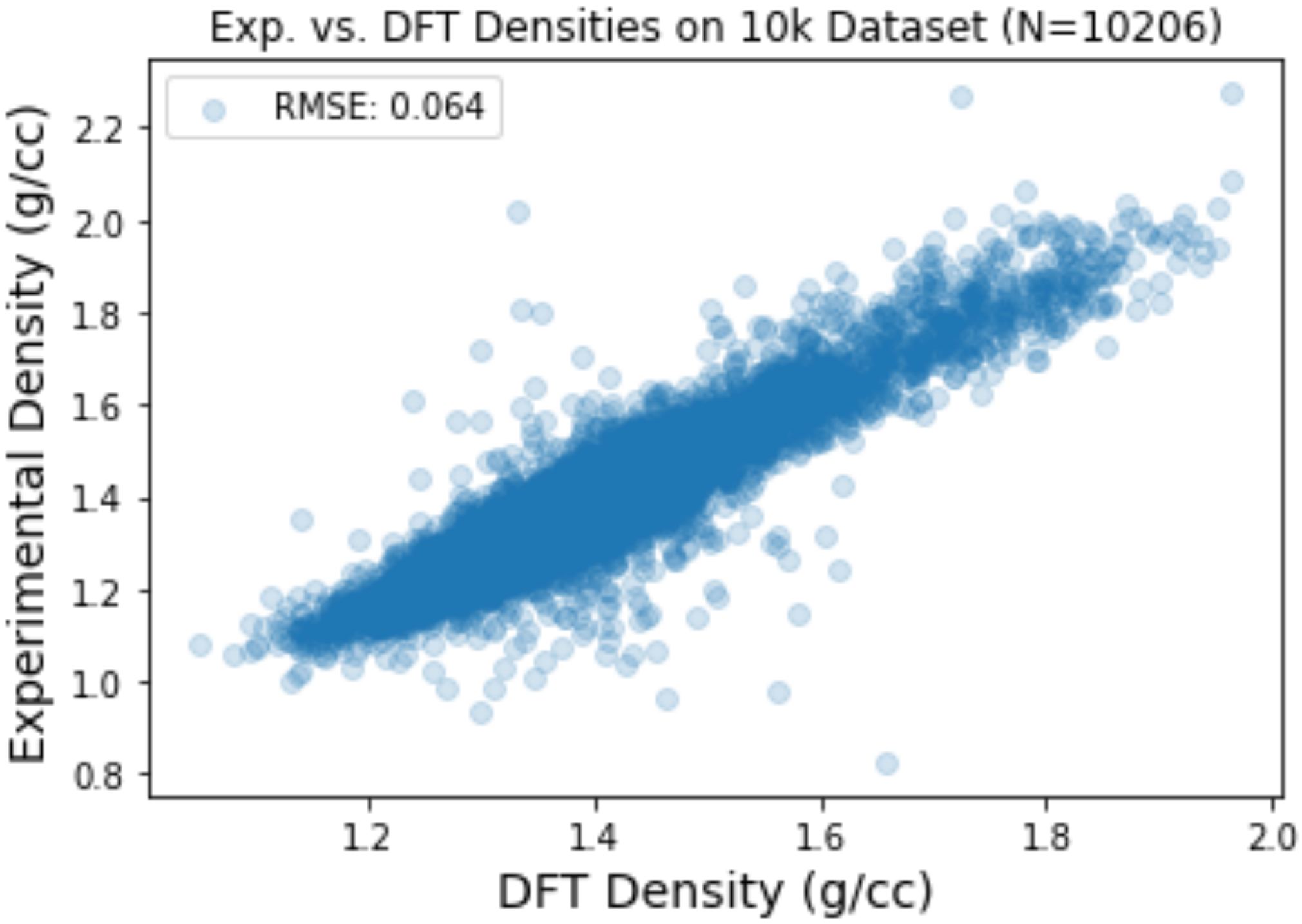

**Figure S1.** Comparison of DFT-calculated densities from our high-throughput DFT pipeline against experimentally determined densities, obtained from the CCDC database.

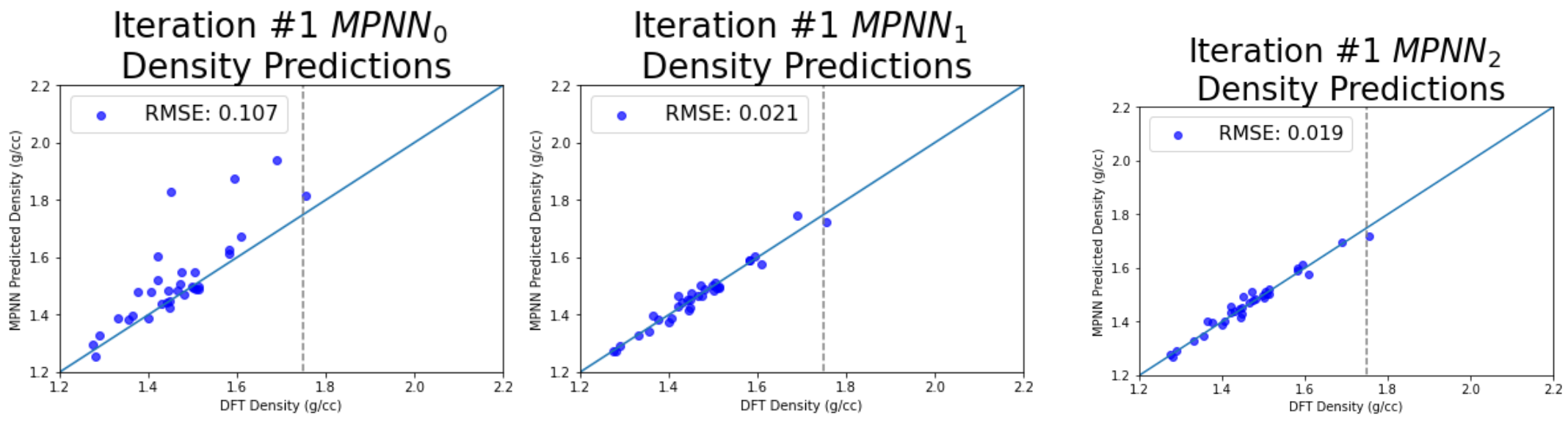

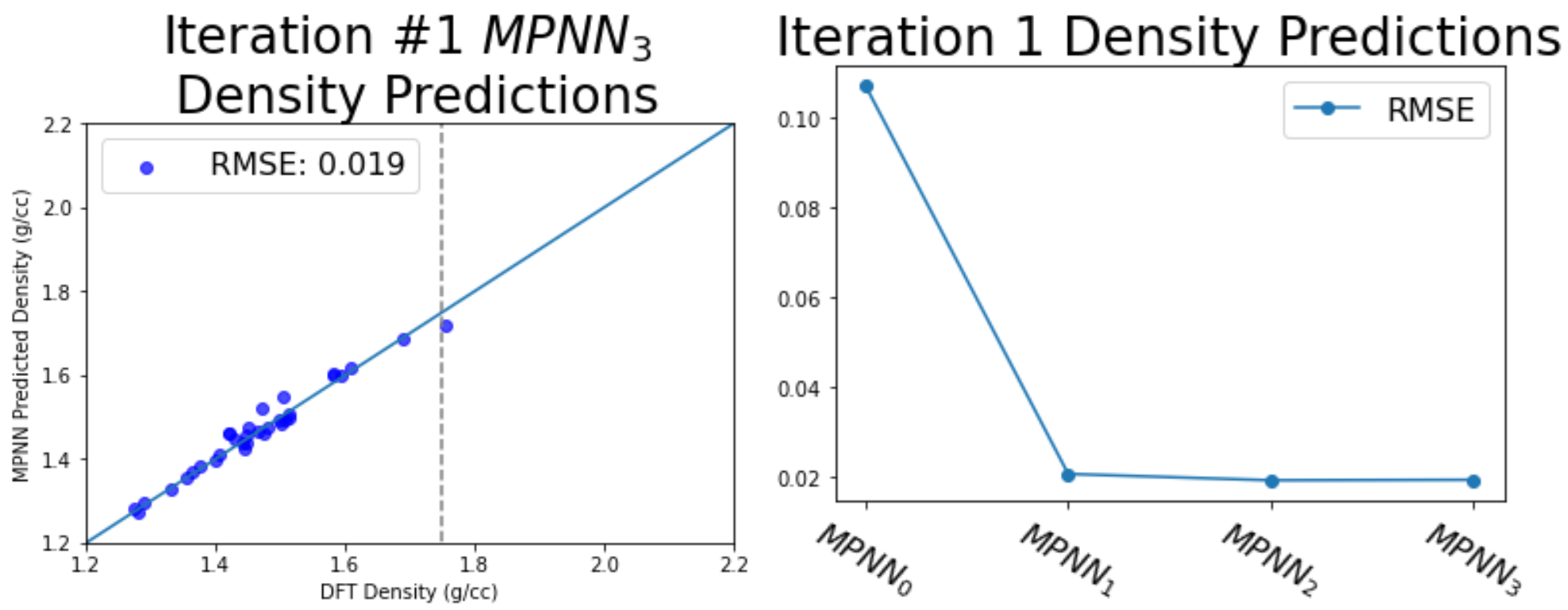

**Figure S2.** Performance of MPNNs at predicting the DFT-calculated density values of a 10% hold-out test set of molecules from the first active learning iteration. In all cases, the test set molecules are not seen in training. For this figure, we do not report precision or recall since there is only a single high-density molecule in the first active learning iteration.

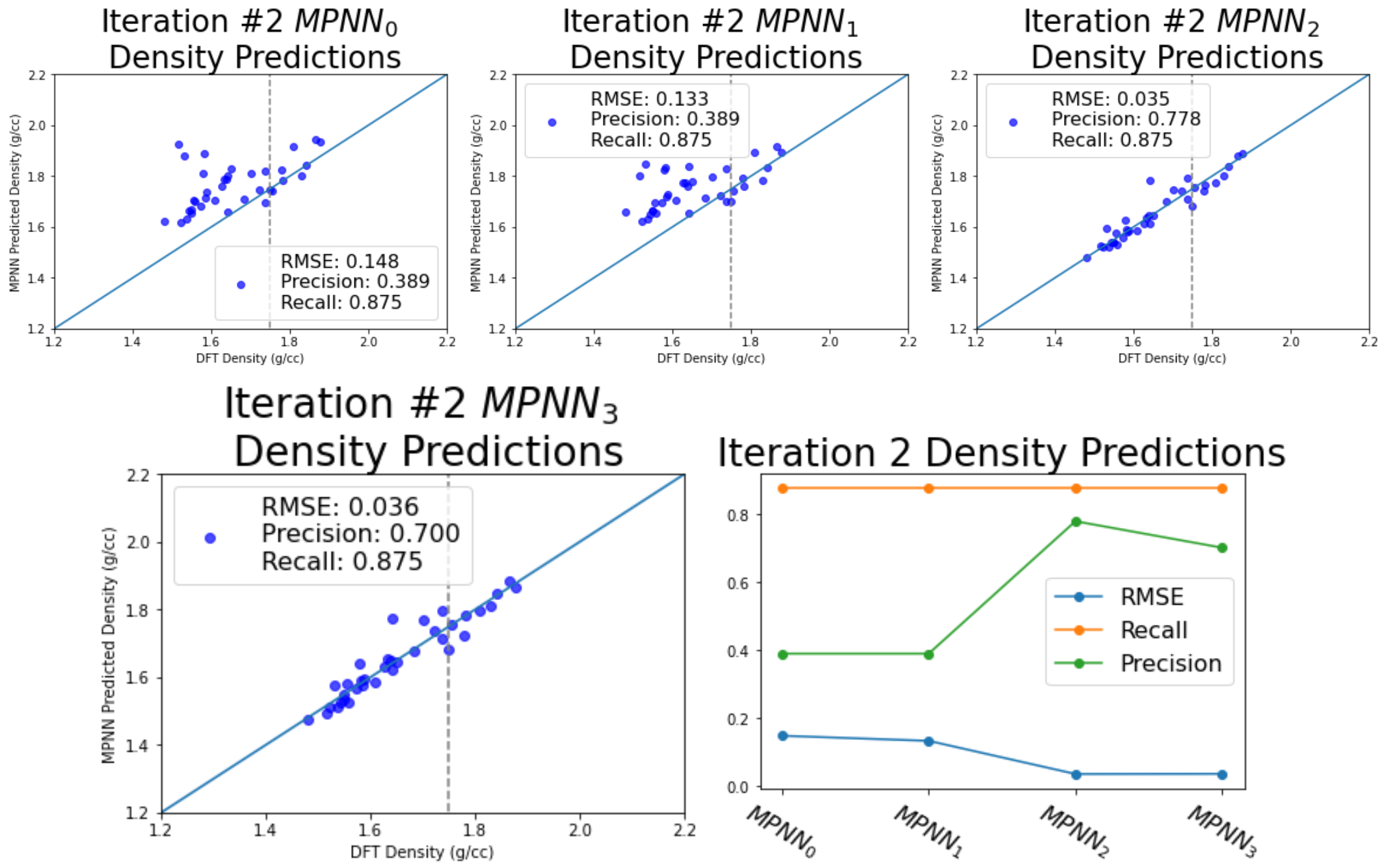

**Figure S3.** Performance of MPNNs at predicting the DFT-calculated density values of a 10% hold-out test set of molecules from the second active learning iteration. In all cases, the test set molecules are not seen in training.

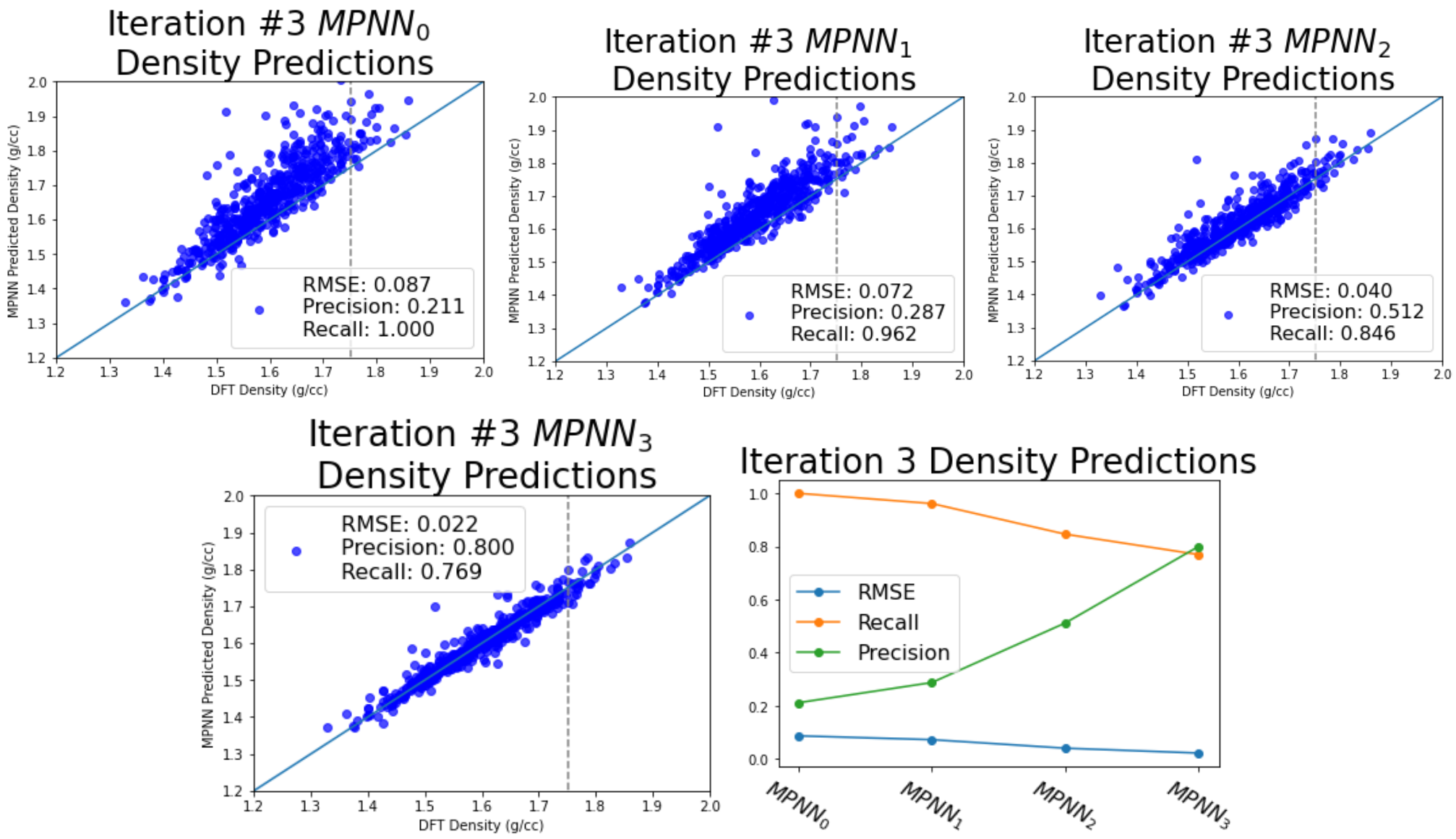

**Figure S4.** Performance of MPNNs at predicting the DFT-calculated density values of a 10% hold-out test set of molecules from the third active learning iteration. In all cases, the test set molecules are not seen in training. The performance statistics are summarized in the bottom right.

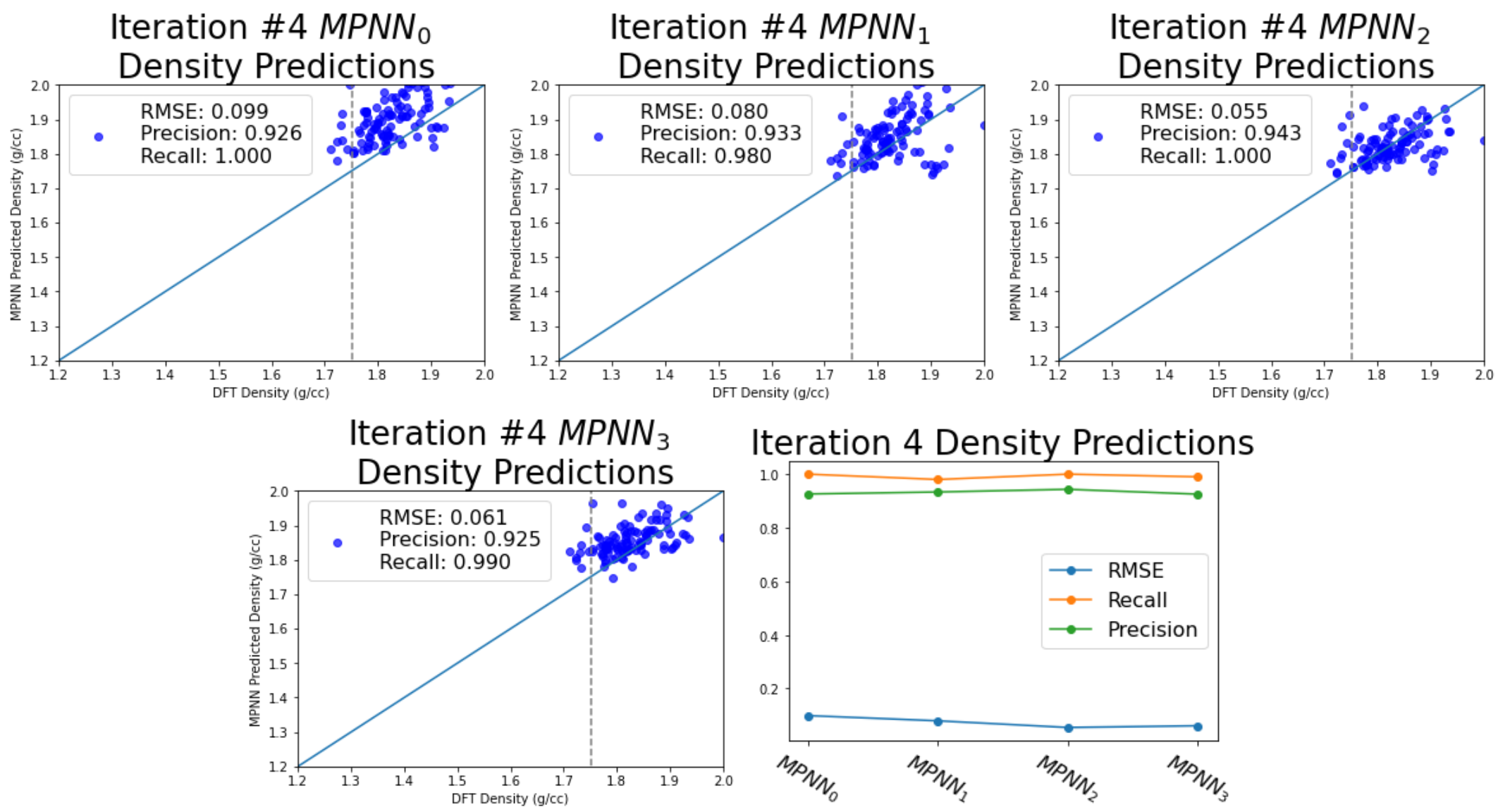

**Figure S5**. Performance of MPNNs at predicting the DFT-calculated density values of all molecules from the fourth active learning iteration. In all cases, the test set molecules are not seen in training. The performance statistics are summarized in the bottom right.

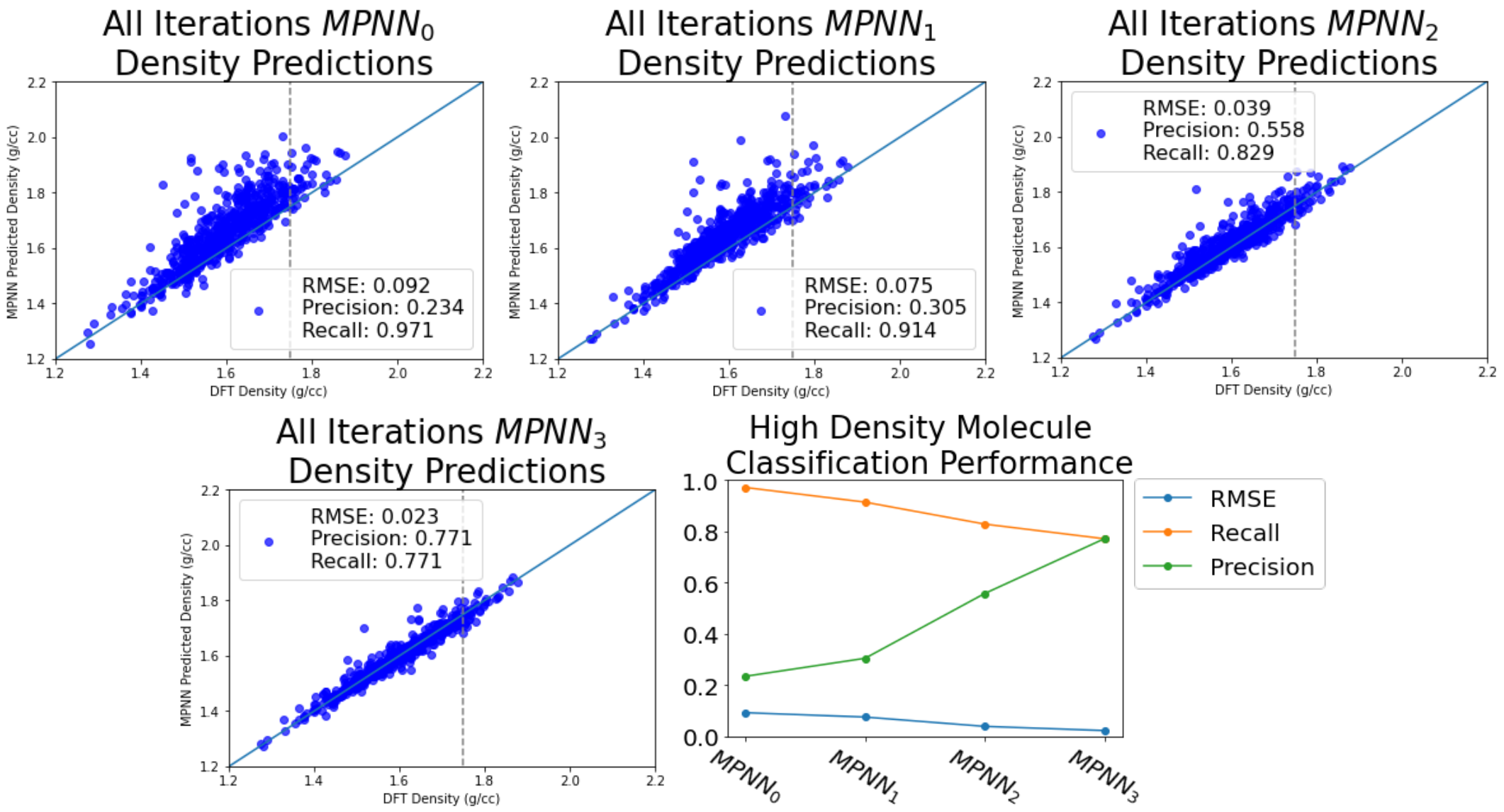

**Figure S6.** Performance of MPNNs at predicting the DFT-calculated density values of molecules drawn from the 10% hold out sets in the first three active learning iterations and the entire fourth iteration. In all cases, the test set molecules are not seen in training.

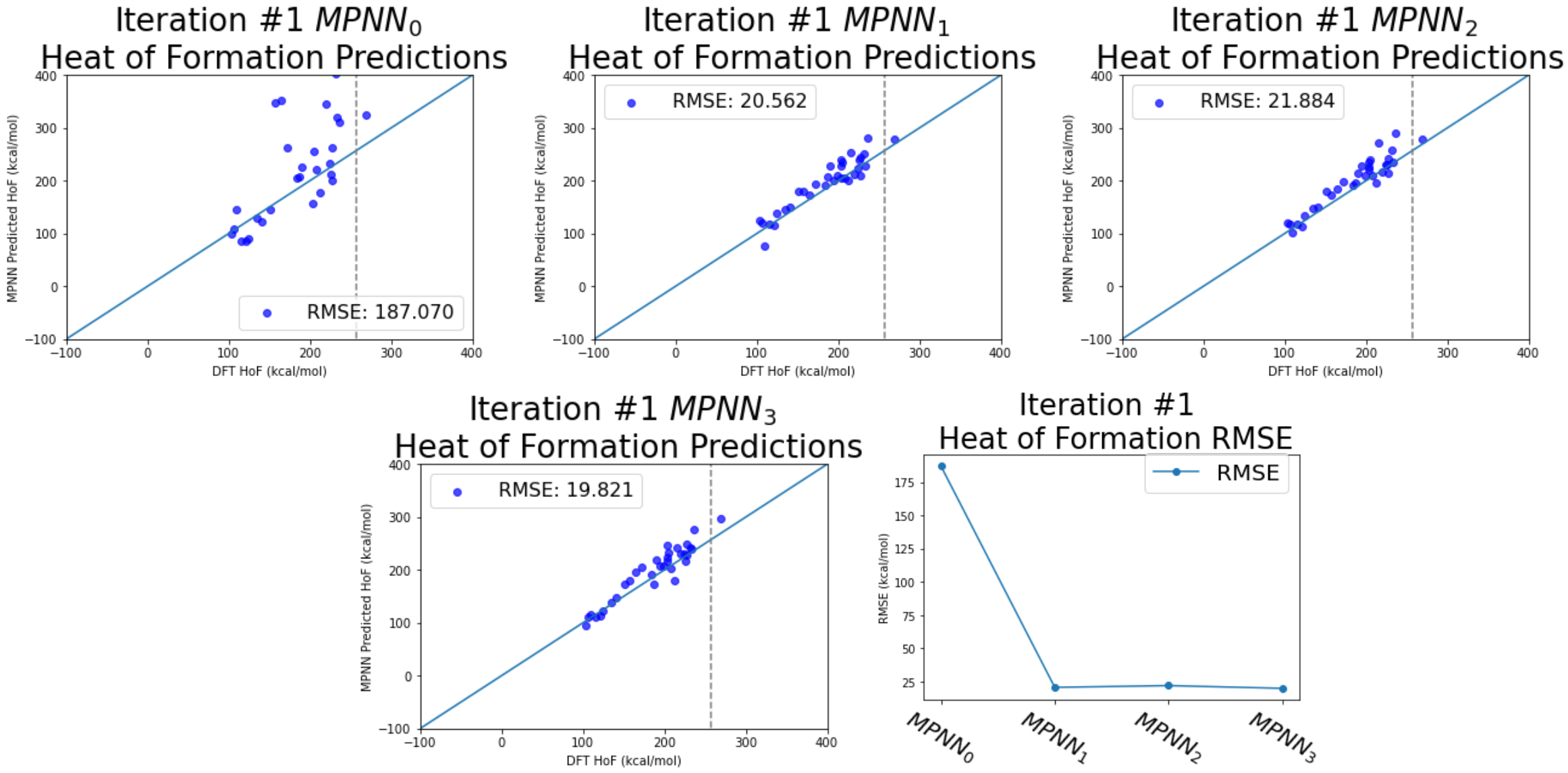

**Figure S7.** Performance of MPNNs at predicting the DFT-calculated solid heat of formation values of a 10% hold-out test set of molecules from the first active learning iteration. In all cases, the test set molecules are not seen in training. The performance statistics are summarized in the bottom right. For this figure, we do not report precision or recall since there is only a single high heat of formation molecule in the first active learning iteration.

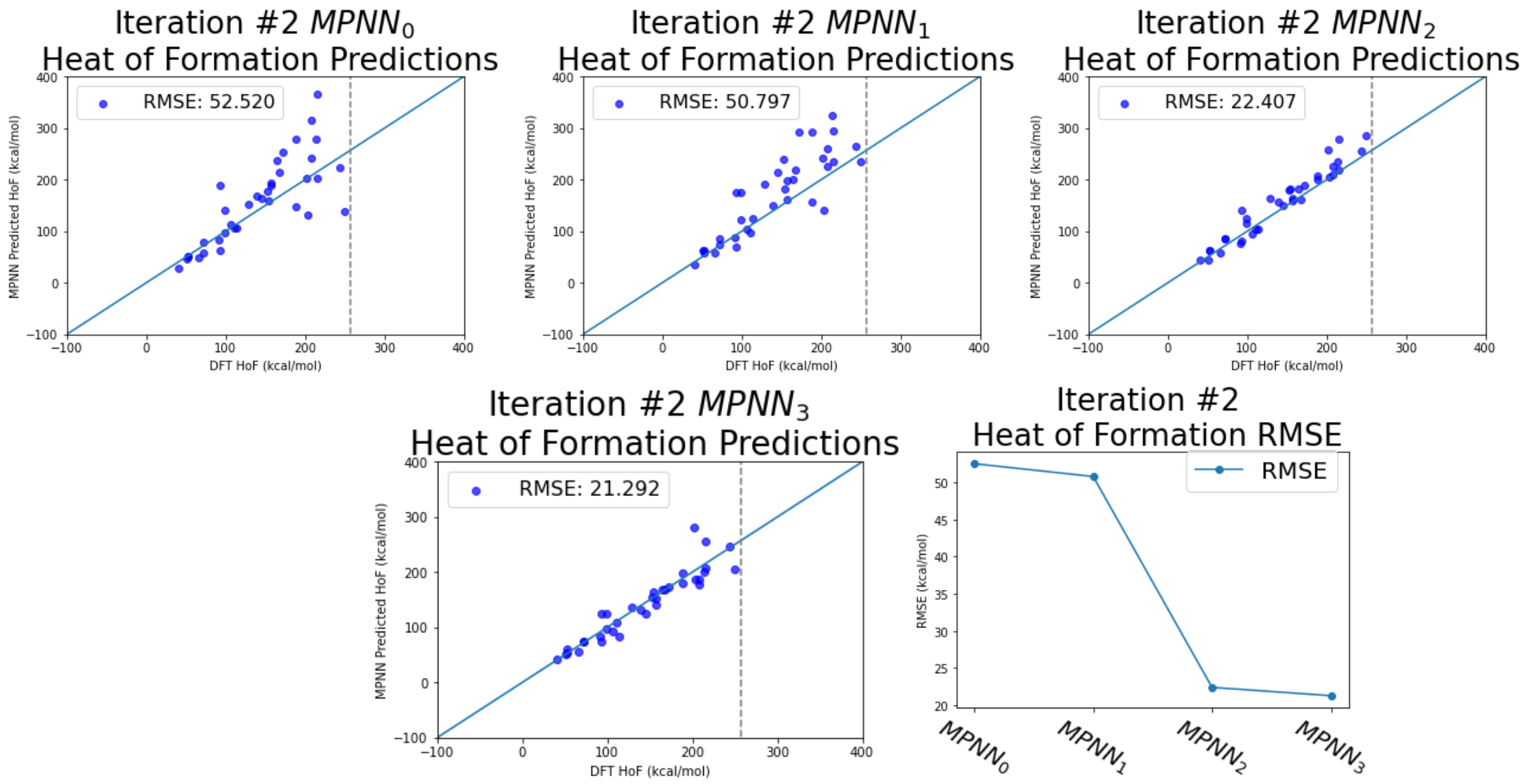

**Figure S8.** Performance of MPNNs at predicting the DFT-calculated solid heat of formation values of a 10% hold-out test set of molecules from the second active learning iteration. In all cases, the test set molecules are not seen in training. The performance statistics are summarized in the bottom right. For this figure, we do not report precision or recall since there are no high heat of formation molecules in the second active learning iteration.

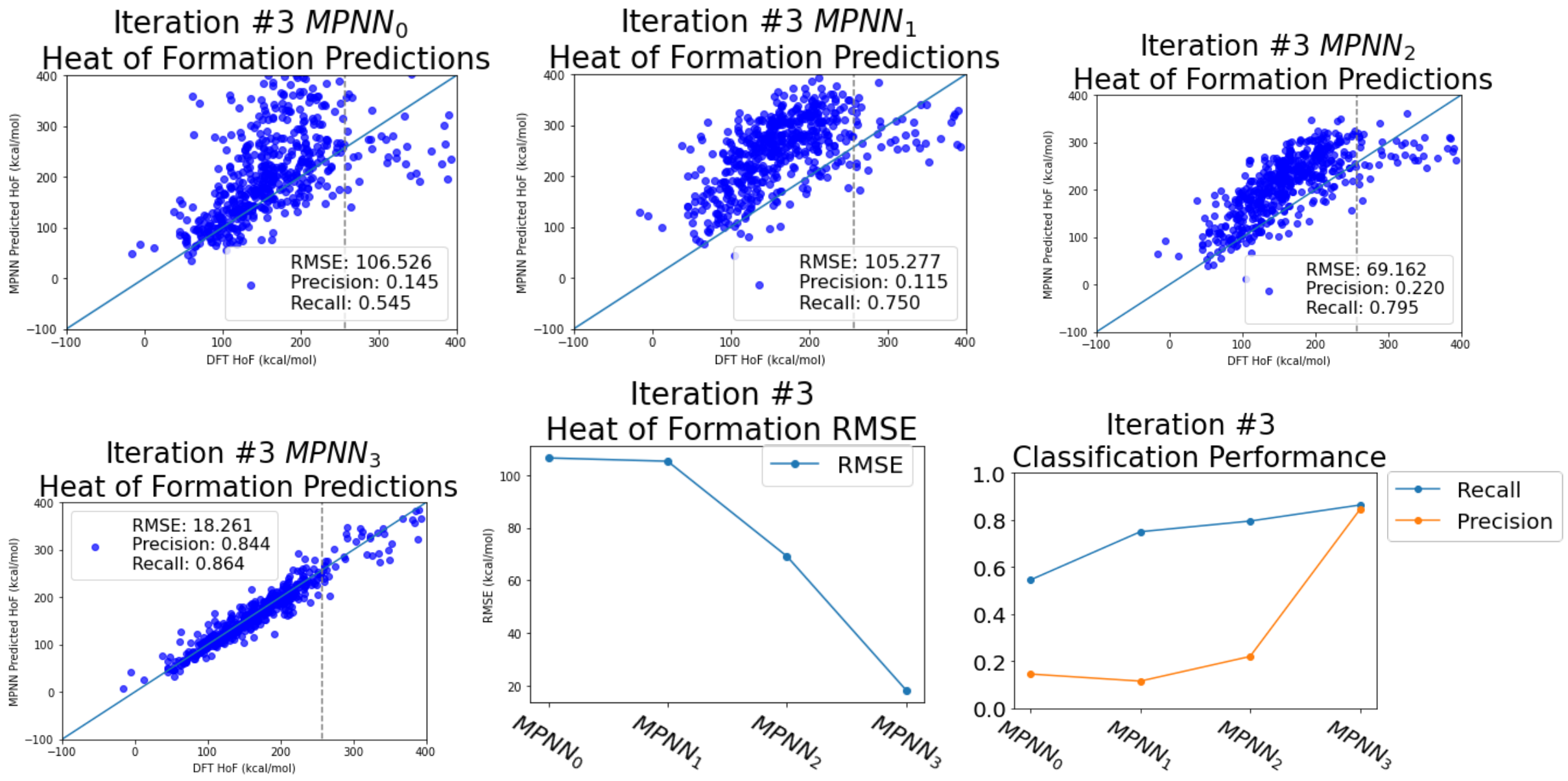

**Figure S9.** Performance of MPNNs at predicting the DFT-calculated solid heat of formation values of a 10% hold-out test set of molecules from the third active learning iteration. In all cases, the test set molecules are not seen in training.

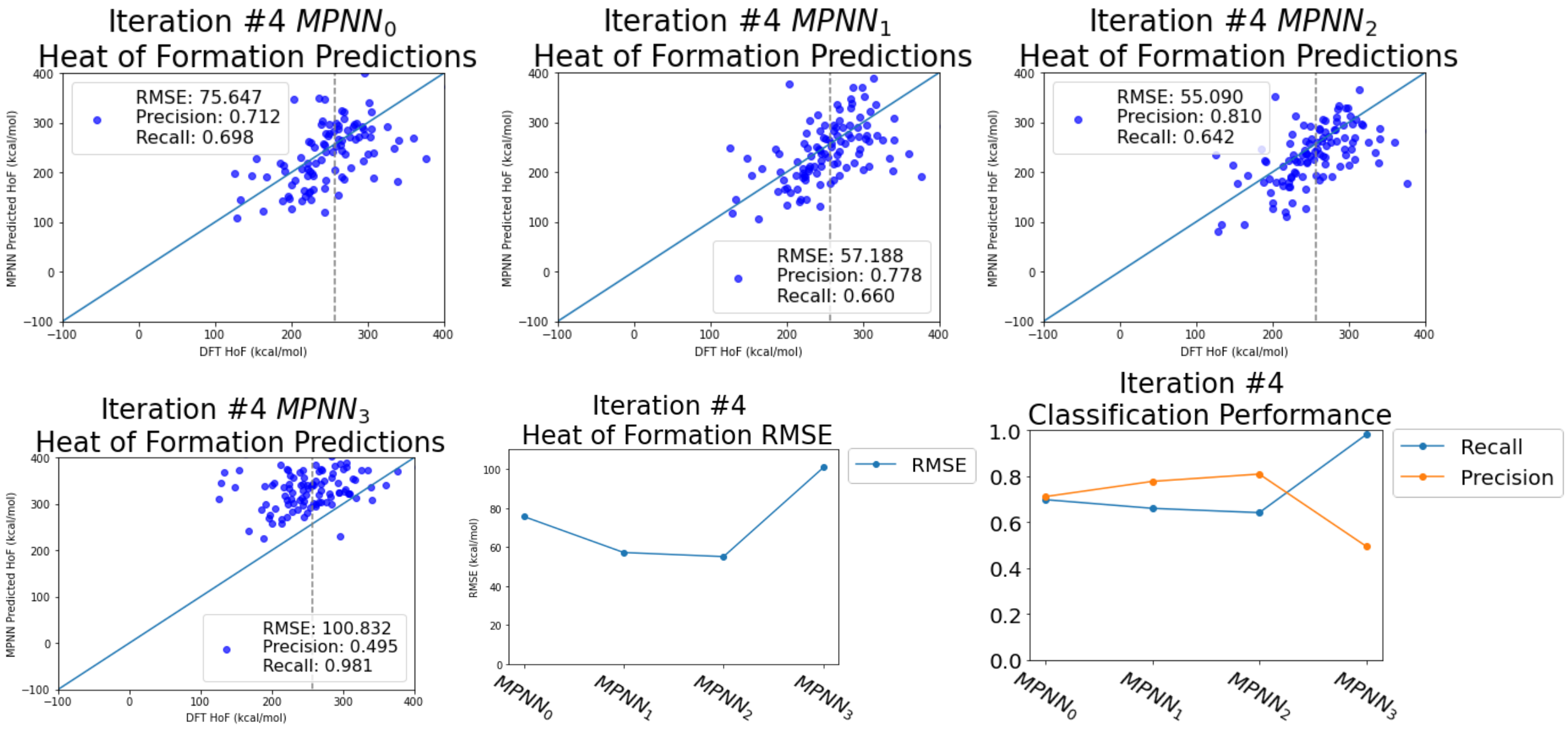

**Figure S10.** Performance of MPNNs at predicting the DFT-calculated solid heat of formation values of all molecules from the fourth active learning iteration. In all cases, the test set molecules are not seen in training.

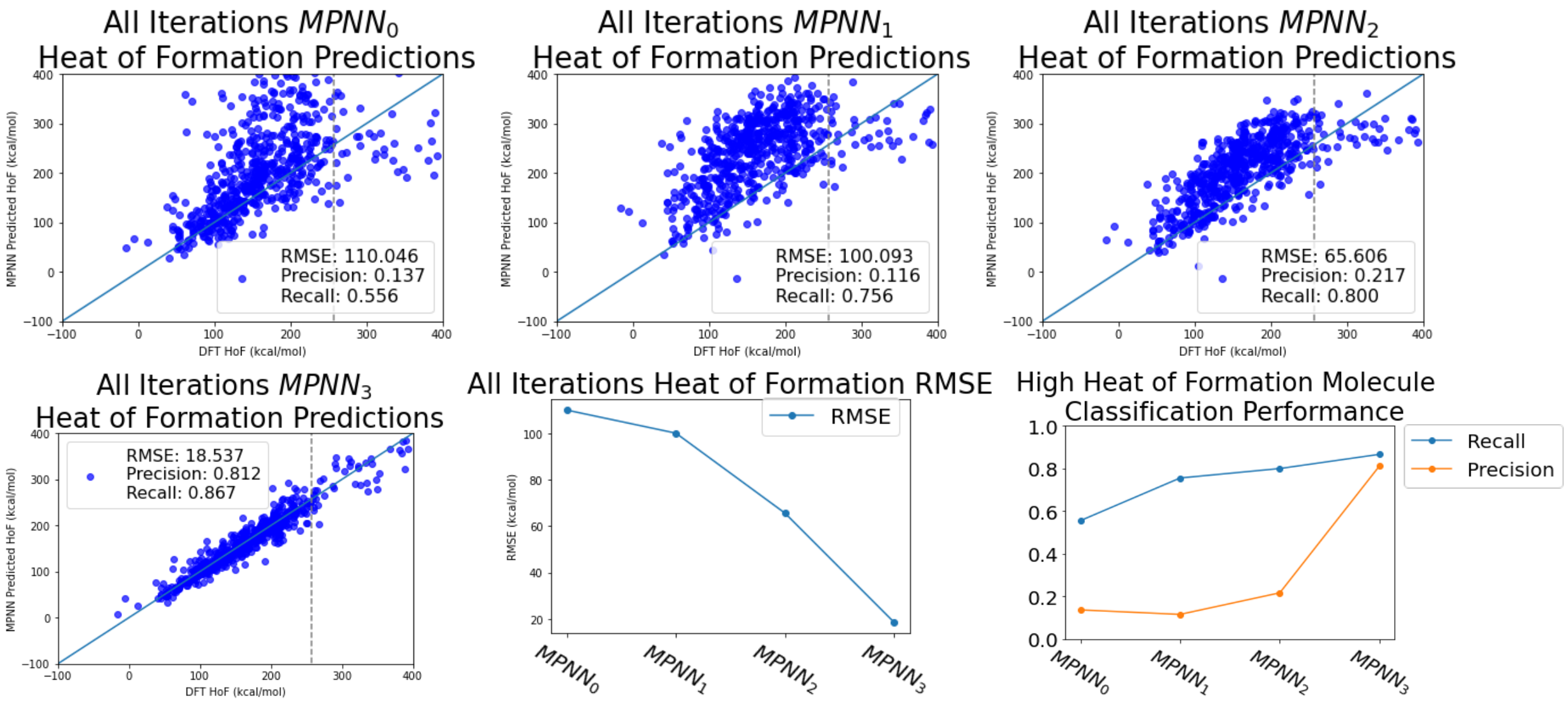

**Figure S11.** Performance of MPNNs at predicting the DFT-calculated solid heat of formation values of molecules drawn from the 10% hold out sets in the first three active learning iterations and the entire fourth iteration. In all cases, the test set molecules are not seen in training.

**Supplementary Note 1**

To supplement the results in the main text showing that the $MPNN_0$ does not extrapolate well to predict the properties of the generated molecules, we consider four alternative approaches for improving the generalization capabilities.

1.) **MPNN 10k Ensemble:** First, we try a deep ensemble approach whereby we independently train 5 MPNN models on the 10k Dataset with different random seeds. The predictions are then obtained by ensembling these five independently trained models. This approach was motivated by the large body of prior work showing that deep ensembles can improve both predictive accuracy and robustness to dataset shift.[1]

2.) **MPNN (All CCDC):** Second, we explore if the generalization capabilities of the MPNN can be improved by increasing the diversity of the training data. For this experiment, rather than only training on the 10k Dataset, we train the density MPNN model on 290,300 experimentally measured densities in the CCDC dataset.

3.) **MoLFormer:** Thirdly, we explore if using chemical foundation models in place of the MPNN as a property predictor leads to improved generalization performance.[2] In particular, we benchmark the MoLFormer foundation model, which was pretrained on a diverse set of 1.1 billion molecules.[2]

4.) **$MPNN_3$ (active learning):** Finally, we compare these approaches to our active learning approach trained on three iterations of active learning ($MPNN_3$), whereby we iteratively retrain the MPNN property predictors on the DFT-calculated properties of the generated molecules.

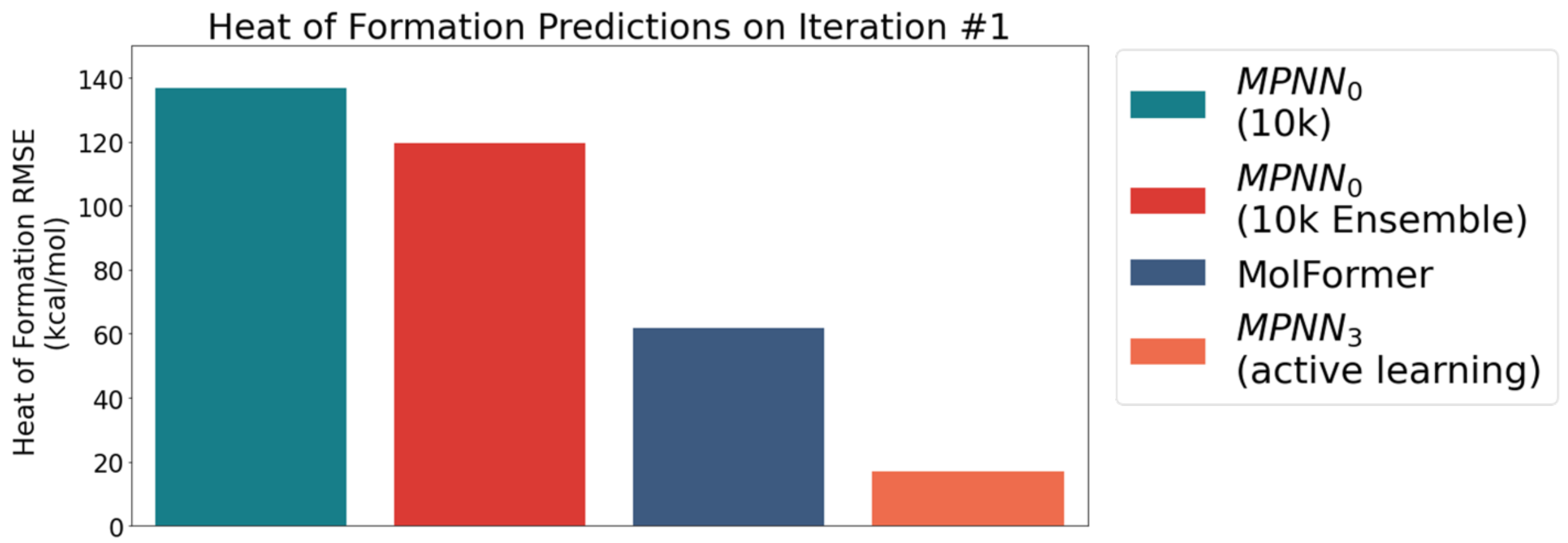

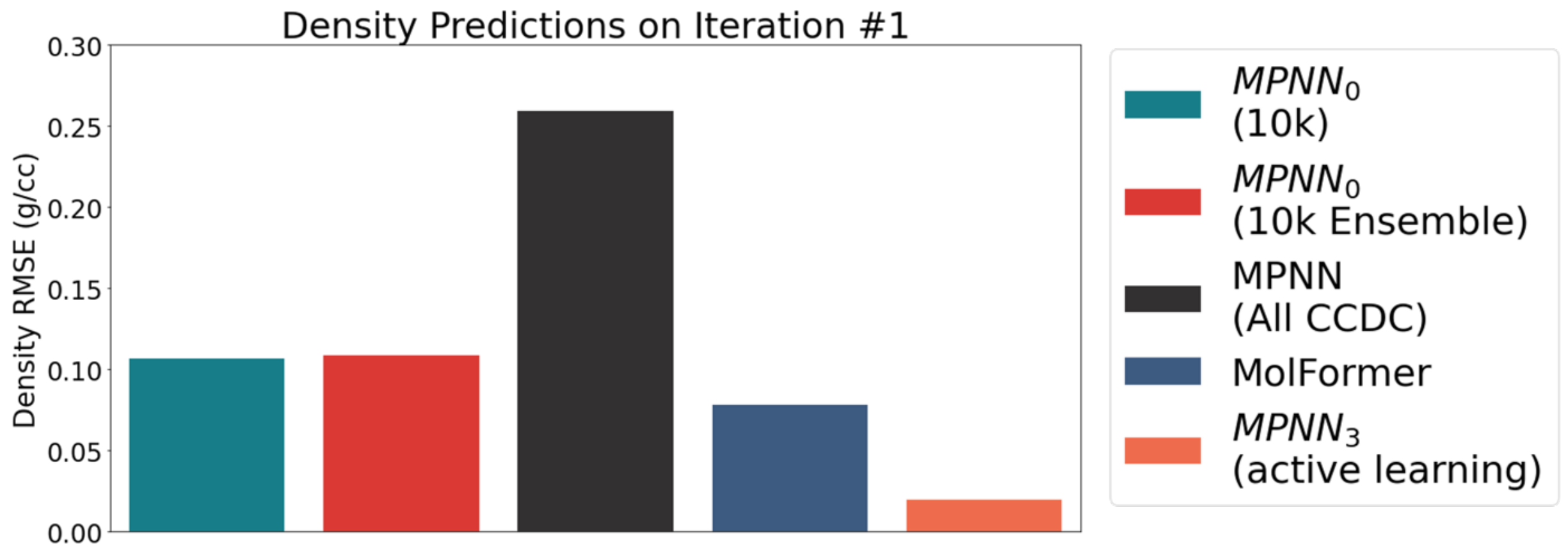

**Figure S12.** **a)** Various models' root mean square error (RMSE) prediction performance for (a) heat of formation and (b) density on a hold-out test set (10%) of the generated molecules in the first iteration (Table 3). The models explored are a single Chemprop MPNN trained on the 10k dataset (green), an ensemble of five Chemprop MPNN models trained on the 10k dataset with different random seeds (red), a single Chemprop MPNN model trained on all 290,300 CCDC experimental densities (black), the MolFormer foundation model fine-tuned on the 10k density dataset (blue), and the active-learning retrained MPNN model that has been retrained on the first three iterations, excluding the test set (orange). **b)** MPNN prediction performance on the generated molecules from the first three iterations. For the purposes of this comparison, we prevent data leakage by training MPNNs on 80% of the generated molecules, using 10% for validation, and holding out 10% of the molecules in each batch for testing. All other models are tested on the same collection of 10% test sets of molecules. The reported errors correspond to average test set prediction performance across the five splits.

**Combining Retrained MPNNs with REINVENT**

In the main text, we show that iteratively retraining the MPNN property prediction models in an active-learning fashion improves the generative performance of JANUS to exceed that of REINVENT. Here, we explore if using the $MPNN_3$ model (retrained over 3 iterations of active learning) improves the performance of REINVENT relative to using the $MPNN_0$ model with REINVENT. For this experiment, we generate 500 molecules with REINVENT using the same experimental setting described in the main text Methods section, except where molecular $\Delta H_{f,s}$ and $\rho$ values are obtained from the $MPNN_3$ models.

**Table S1.** Comparison of REINVENT generative performance when using $MPNN_0$ and $MPNN_3$ models for property prediction.

| Approach | Valid | Top DFT $\rho$ (g/cc) | Top DFT $\Delta H_{f,s}$ (kcal/mol) | Mean DFT $\rho$ (g/cc) | Mean DFT $\Delta H_{f,s}$ (kcal/mol) | % DFT Stable Molecules[b] |
|---|---|---|---|---|---|---|
| REINVENT w/ $MPNN_0$ | **1.00** | 1.901 | **417** | 1.857 | 296 | **6% (30/500)** |
| REINVENT w/ $MPNN_3$ | **1.00** | **1.905** | **417** | **1.870** | **319** | 4% (20/500) |

We note that this ablation experiment should not be considered as "active learning" since in both experimental settings, REINVENT is only performing "one-shot" generation of molecules and is not being retrained on the outputs of the DFT calculations. Rather, Table S1 explores whether the iterative retraining of the MPNN property prediction model can improve the performance of other generative models, beyond the one directly used in the active learning loop. We find that combining $MPNN_3$ with REINVENT significantly improves the average properties of generated molecules, but does not significantly improve the extrapolation performance of the REINVENT model. This finding helps to underscore the main conclusion of our work that it is only through multiple rounds of iterative retraining (i.e. active learning) that the extrapolation performance of the generative model is improved.

**Visual Comparison of 10k Dataset molecules and Generated Molecules**

To obtain an intuitive understanding of how the molecules generated by our active learning pipeline differ from the known molecules in the 10k Dataset, we first featurize all molecules with the RdKit Descriptors, implemented in the DeepChem package. Then, all features are normalized across all molecules. In Table S2, we illustrate how the generated molecules differ from the 10k Dataset by listing the molecular features that have the largest absolute normalized shift between the 10k Dataset and the generated molecules. In Figure S13, we plot the distributions of the 10 features that have the largest normalized absolute shift between the 10k Dataset and all generated molecules.

**Table S2.** Comparison of the molecules in the 10k Dataset and the four active learning iterations. Feature descriptions are taken directly from the RDKit documentation. To aid in the interpretability of each feature, we depict molecules with extreme feature values. For features which are higher on average among the generated molecules, we depict the generated molecule with the highest feature value and the molecule in the 10k dataset with the lowest feature value. For features which are lower on average among the generated molecules (QED only), we depict the generated molecule with the lowest feature value and the molecule in the 10k dataset with the highest feature value.

| Feature Names | Feat-ure Shift | Feature Description | Example from 10k Dataset | Example from Generated Molecules |
|---|---|---|---|---|
| SMR_VSA3 | 0.359 | MOE MR VSA Descriptor 3 | 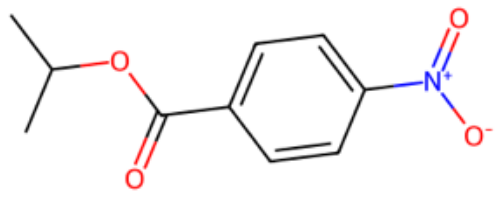 0.00 | 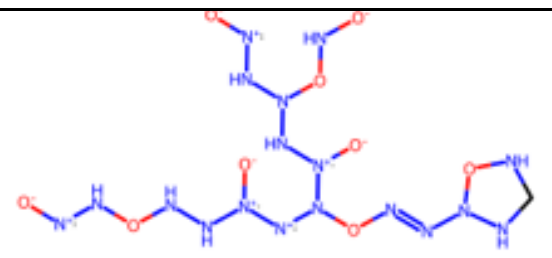 87.45 |
| SlogP_VSA1 | 0.296 | MOE logP VSA Descriptor 1 | 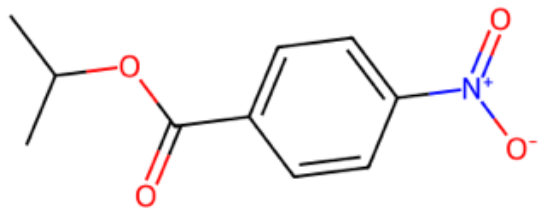 0.00 | 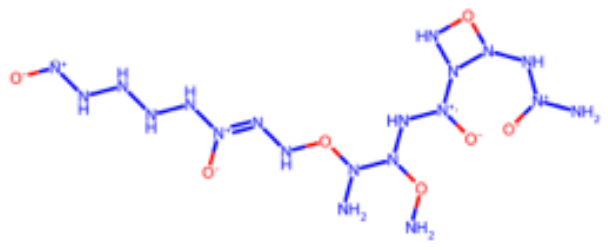 78.12 |
| Qed | 0.285 | Quantitative estimation of drug-likeness | 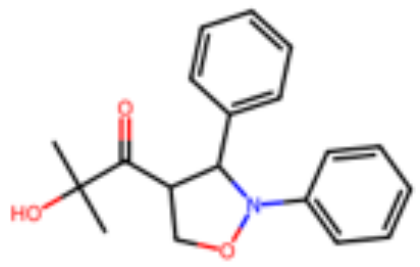 0.942 | 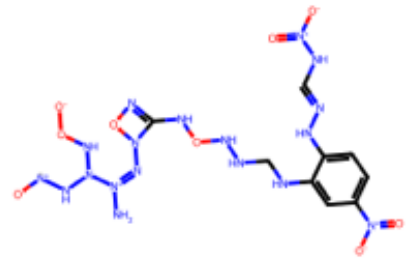 0.00853 |
| fr_NH0 | 0.277 | Number of tertiary amines | 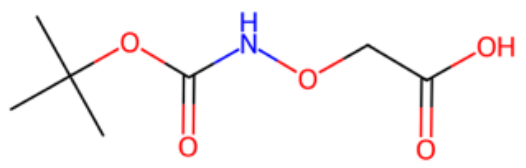 0 | 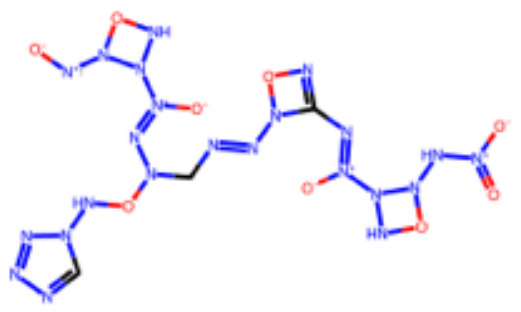 19 |
| NumRadicalElectrons | 0.268 | Number of radical electrons | 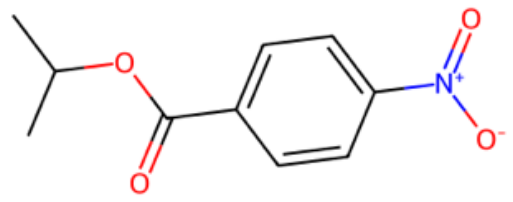 0 | 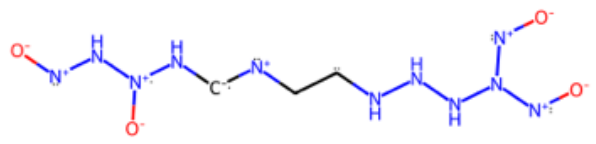 12 |
| fr_hdrzine | 0.230 | Number of hydrazine groups | 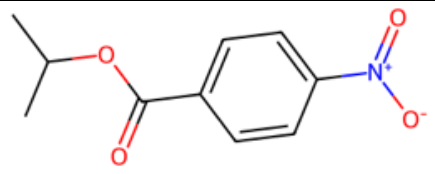 0 | 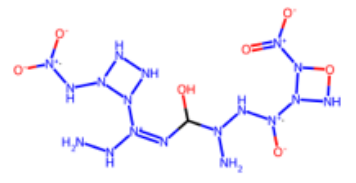 16 |
| MolLogP | 0.222 | Wildman-Crippen LogP Value | 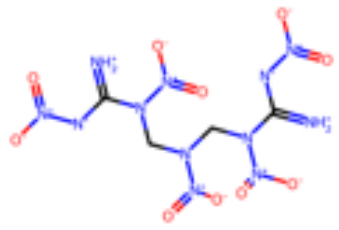 -5.256 | 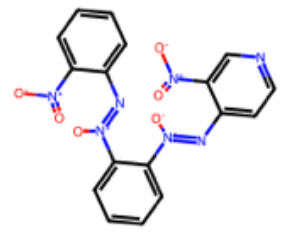 4.749 |
| BCUT2D_MWLOW | 0.221 | BCUT descriptors from J. Chem. Inf. Comput. Sci., Vol. 39, No. 1, 1999. | 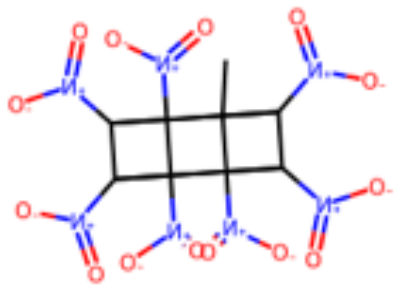 8.799 | 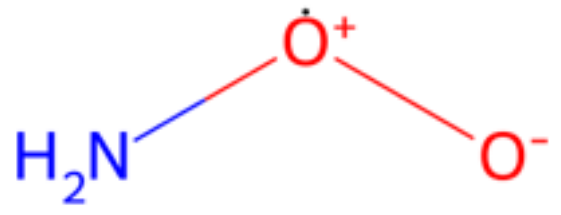 13.524 |
| NumHAcceptors | 0.218 | Number of hydrogen bond acceptors | 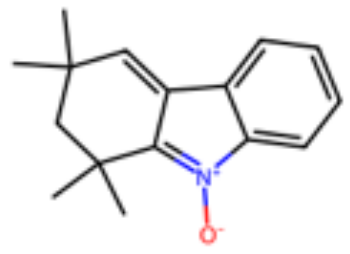 1 | 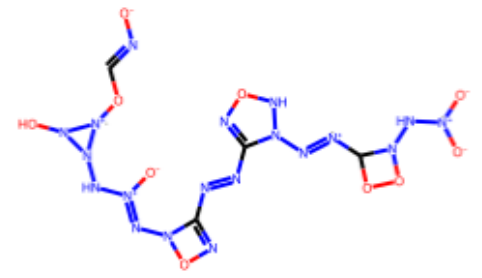 26 |
| NumHeteroatoms | 0.212 | Number of heteroatoms | 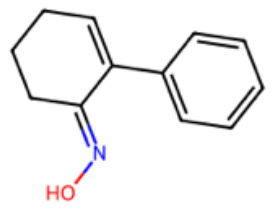 2 | 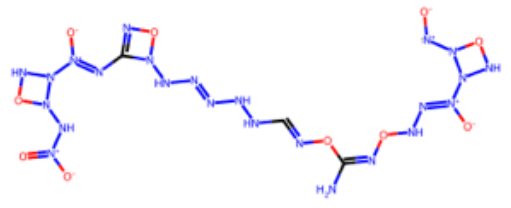 34 |
| NOCount | 0.211 | Number of nitrogens and oxygens | 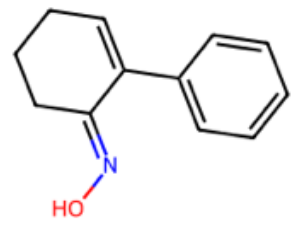 2 | 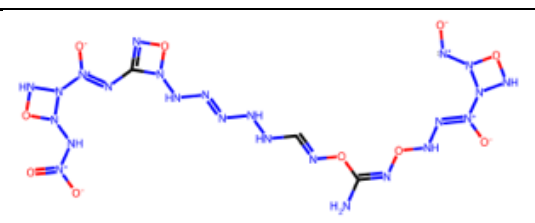 34 |

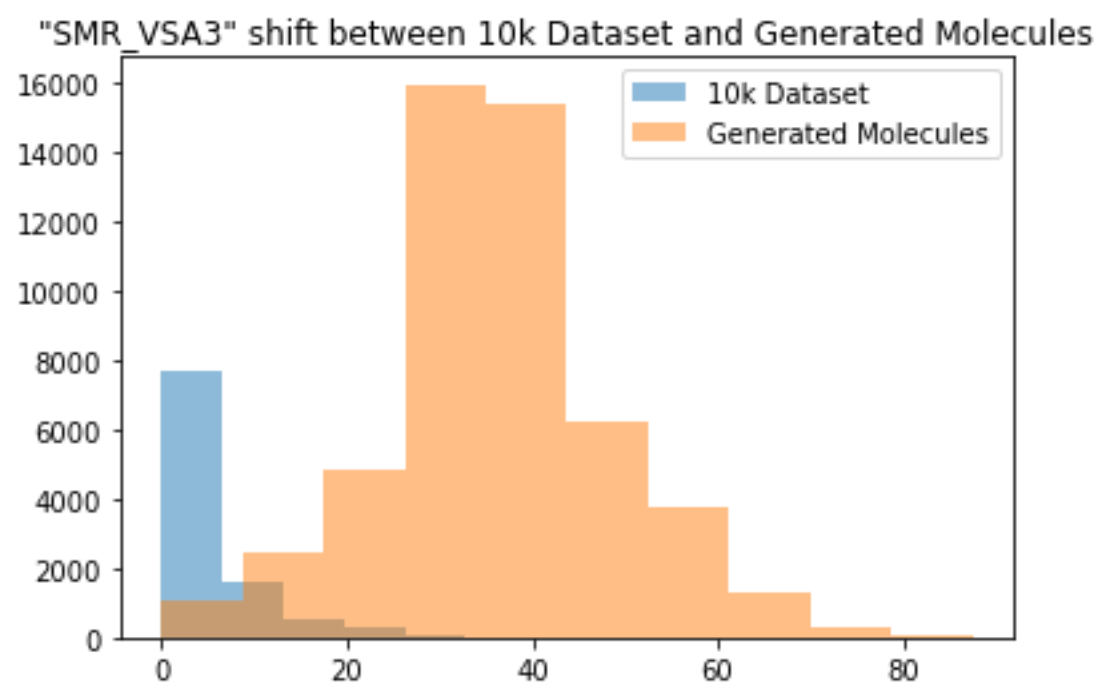

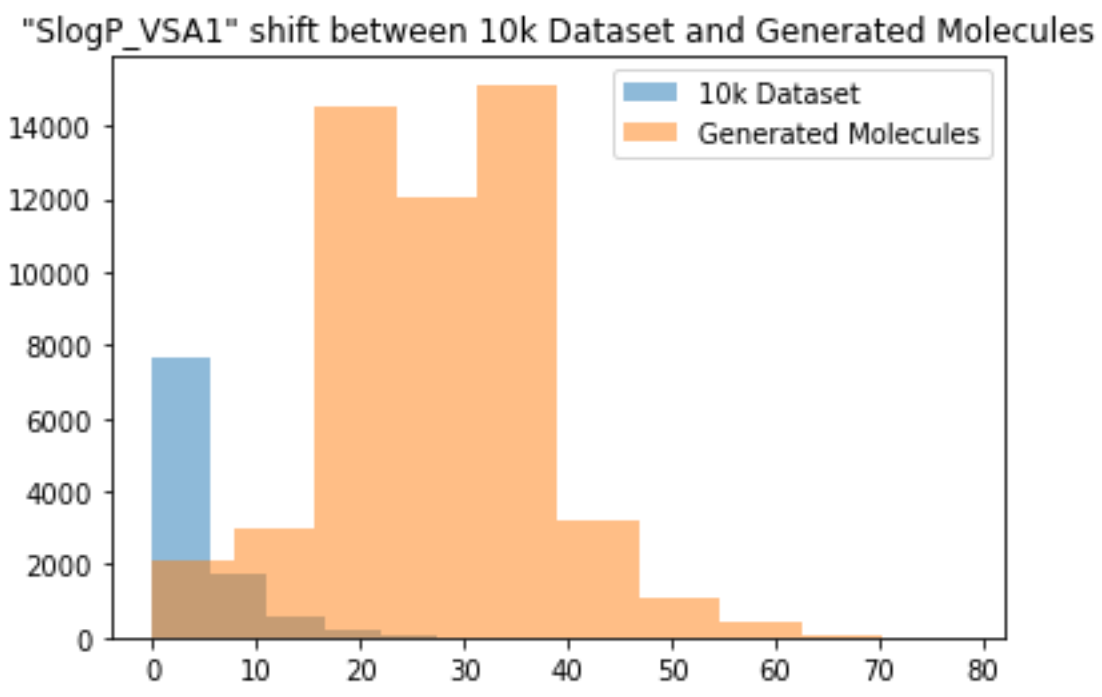

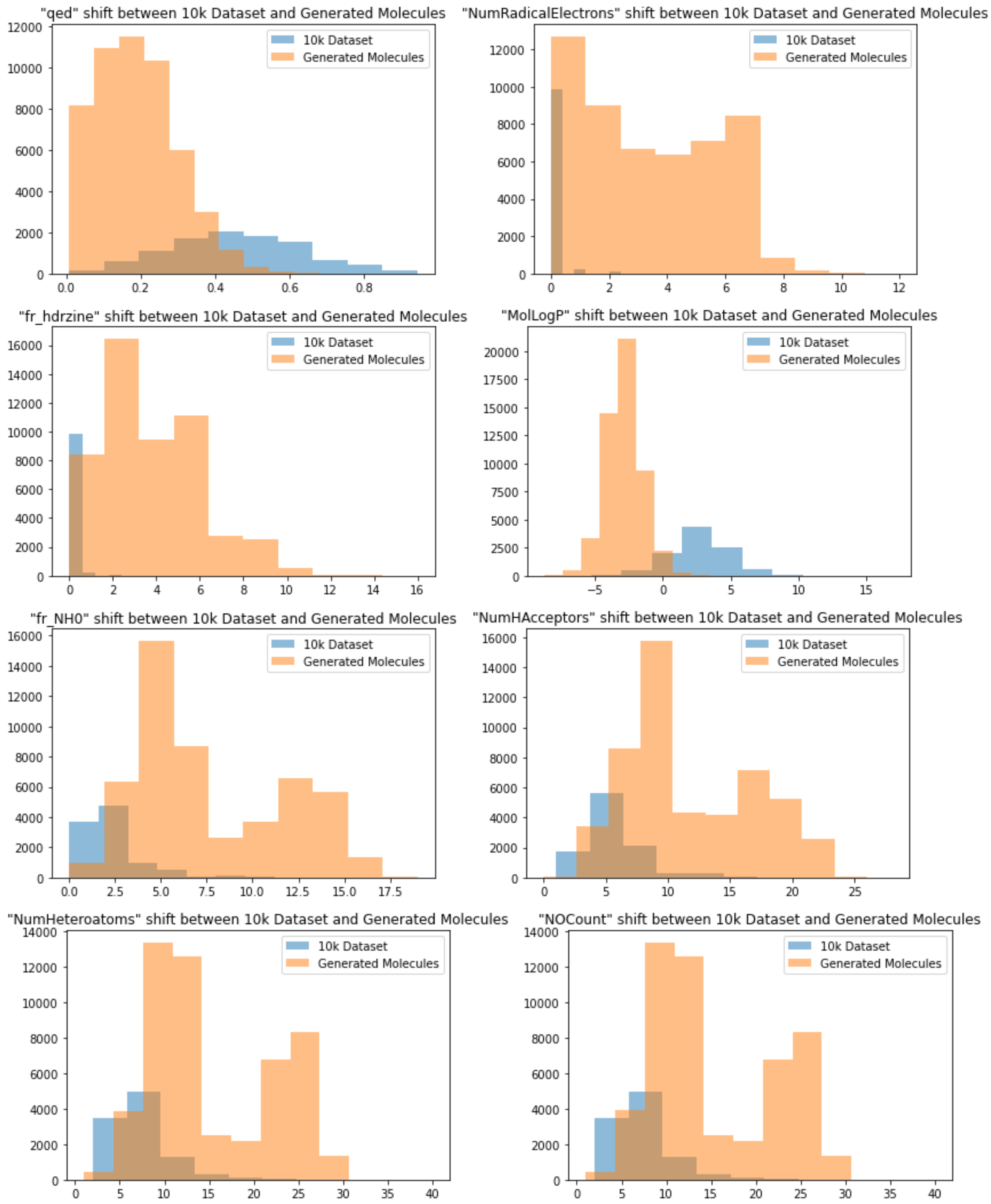

**Figure S13.** Histograms of the 10 RDKit descriptors with the largest normalized absolute shift between the 10k Dataset and all generated molecules.

## Diversity Metrics of Generated Molecules

To further understand the diversity of molecules generated from our active learning pipeline, we calculate a number of diversity metrics from the MOSES molecular generation benchmark.[3] Below, we briefly define each of the diversity metrics in Table S3. We refer interested readers to the original MOSES paper for a full discussion of the diversity metrics. We also note that we explicitly filter out any non-unique generated molecules after each active learning generation.

**Valid:** Fraction of generated molecules with valid SMILES strings, as determined by RDKit's molecular structure parser.
**Frag:** Fragment similarity between the generated molecules and the reference dataset (10k dataset in this case). Frag can range from [0,1], with larger values indicating that the reference dataset and generated molecules have similar molecular fragments. Results are calculated both for a random test set (Test) and a scaffold split test set (TestSF), which is test set containing all molecules with a particular molecular scaffold.
**Scaff:** Similar to the above fragment similarity, except molecular scaffolds are used instead of molecular fragments.
**IntDiv/IntDiv2:** Internal diversity of generated molecules, where diversity is calculated with respect to the Tanimoto similarity of fingerprints of two generated molecules. In IntDiv2, the squared Tanimoto similarity is used instead.
**Novelty:** Fraction of generated molecules not present in the reference dataset (10k dataset).

**Table S3.** Diversity metrics calculated for each iteration of the active learning pipeline.

| Active Learning Iteration | Valid | Frag | | Scaff | | IntDiv | IntDiv2 | Novelty |
|---|---|---|---|---|---|---|---|---|
| | | Test | TestSF | Test | TestSF | | | |
| 1 | 1.0 | 0.035 | 0.117 | 0.0004 | 0.0 | 0.889 | 0.862 | 1.0 |
| 2 | 1.0 | 0.058 | 0.154 | 0.0142 | 0.0 | 0.840 | 0.821 | 1.0 |
| 3 | 1.0 | 0.152 | 0.146 | 0.0081 | 0.0 | 0.879 | 0.866 | 1.0 |
| 4 | 1.0 | 0.0 | 0.0 | 0.0 | 0.0 | 0.853 | 0.818 | 1.0 |

As shown in Table S3, the molecules generated from our approach have very low fragment and scaffold similarity with the 10k dataset, confirming that the generated molecules are indeed extrapolated in chemical space with respect to the training data. At the same time, the generated molecules exhibit high internal diversity, novelty, and validity, indicating that our approach is able to generate a wide range of chemically valid molecules within this extrapolated region of chemical space.

**Synthesizability of Generated Molecules**

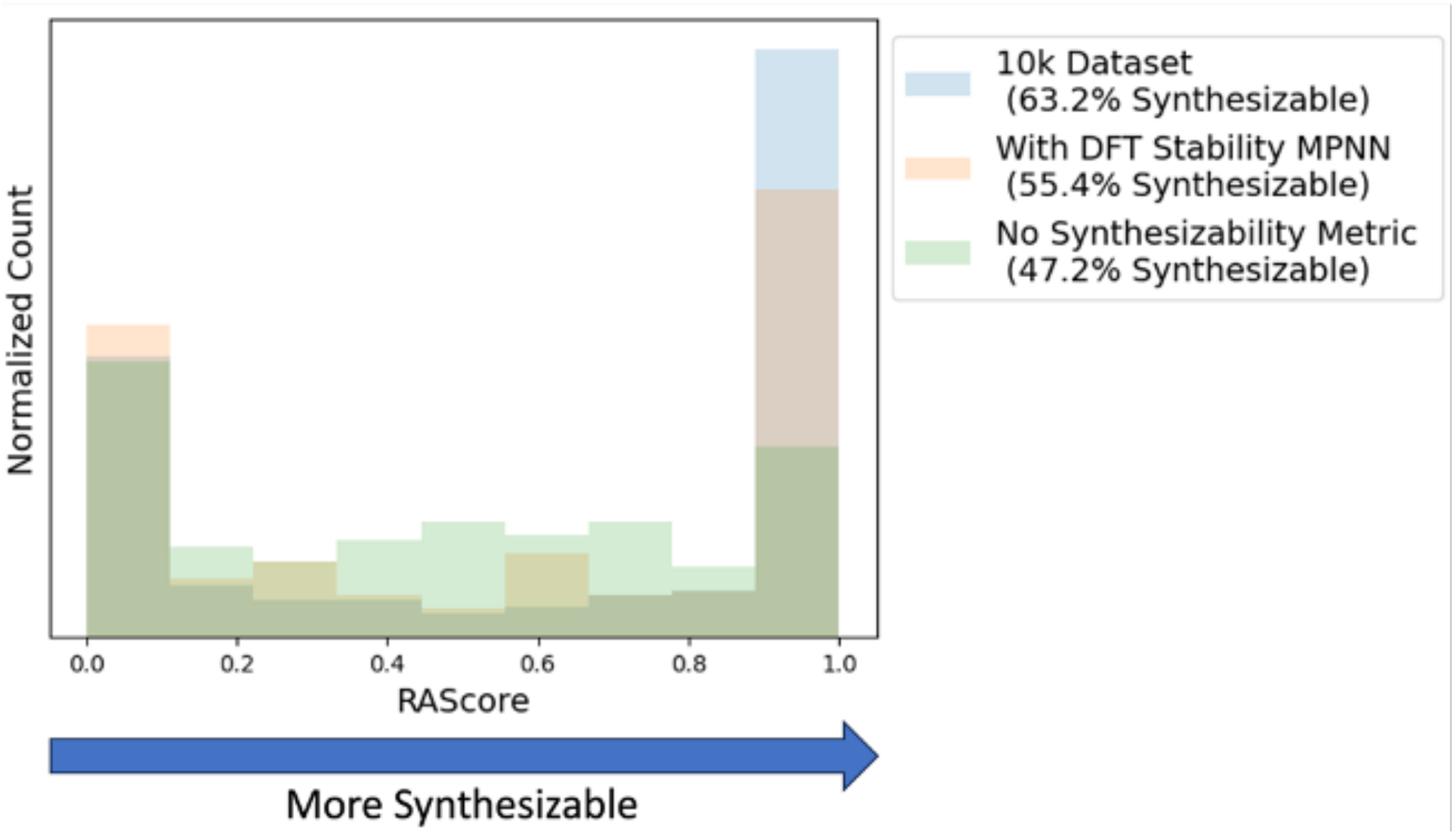

**Figure S14.** Distribution of RAScore values for all molecules in the 10k Dataset (blue), molecules generated by our active learning approach (orange), and molecules generated without any synthesizability criteria as shown in Table 2 (green).

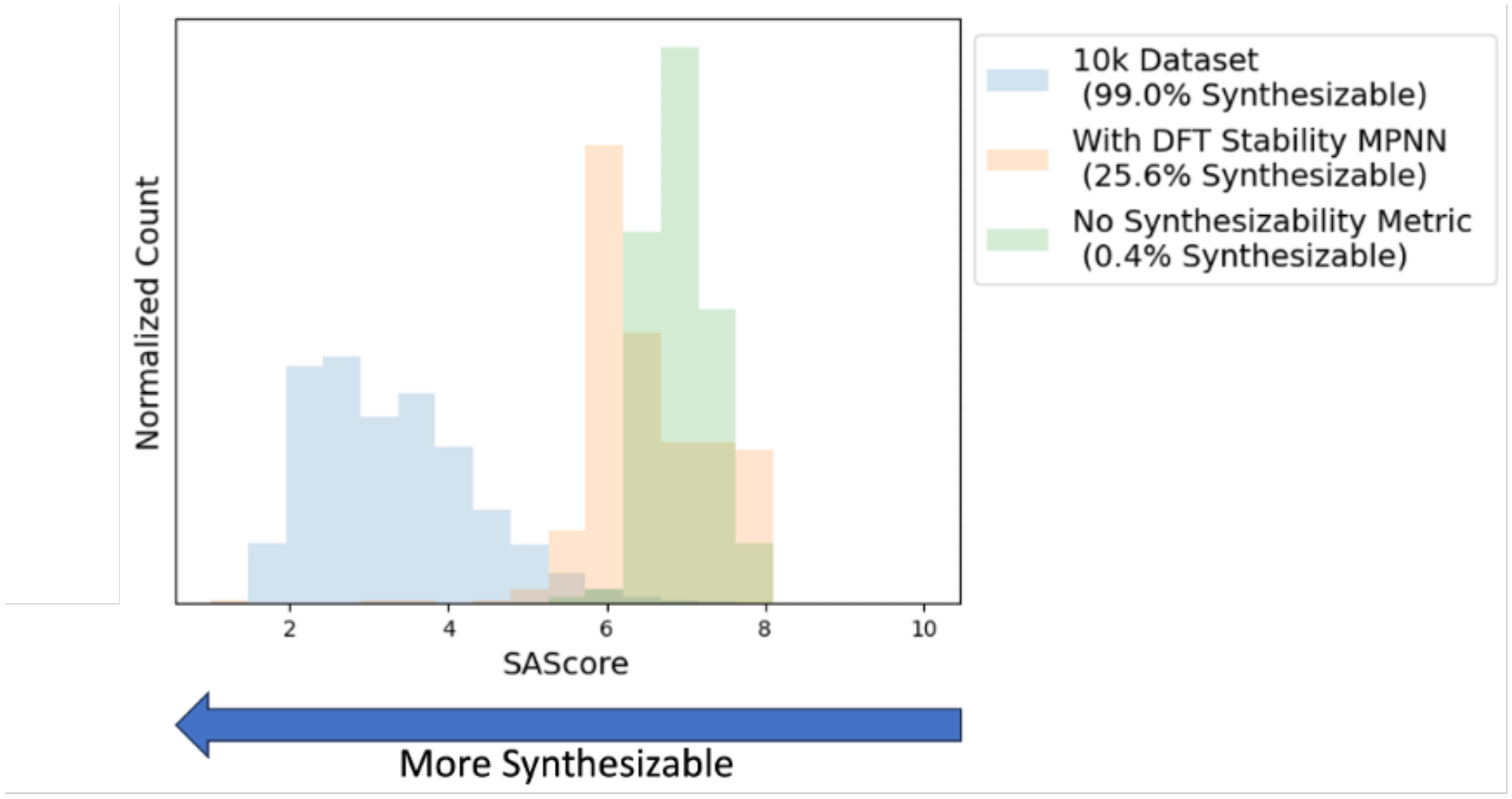

**Figure S15.** Distribution of SAScore values for all molecules in the 10k Dataset (blue), molecules generated by our active learning approach (orange), and molecules generated without any synthesizability criteria as shown in Table 2 (green).

**Comparing MPNN Predictions on Exploration and Exploitation Populations of JANUS**

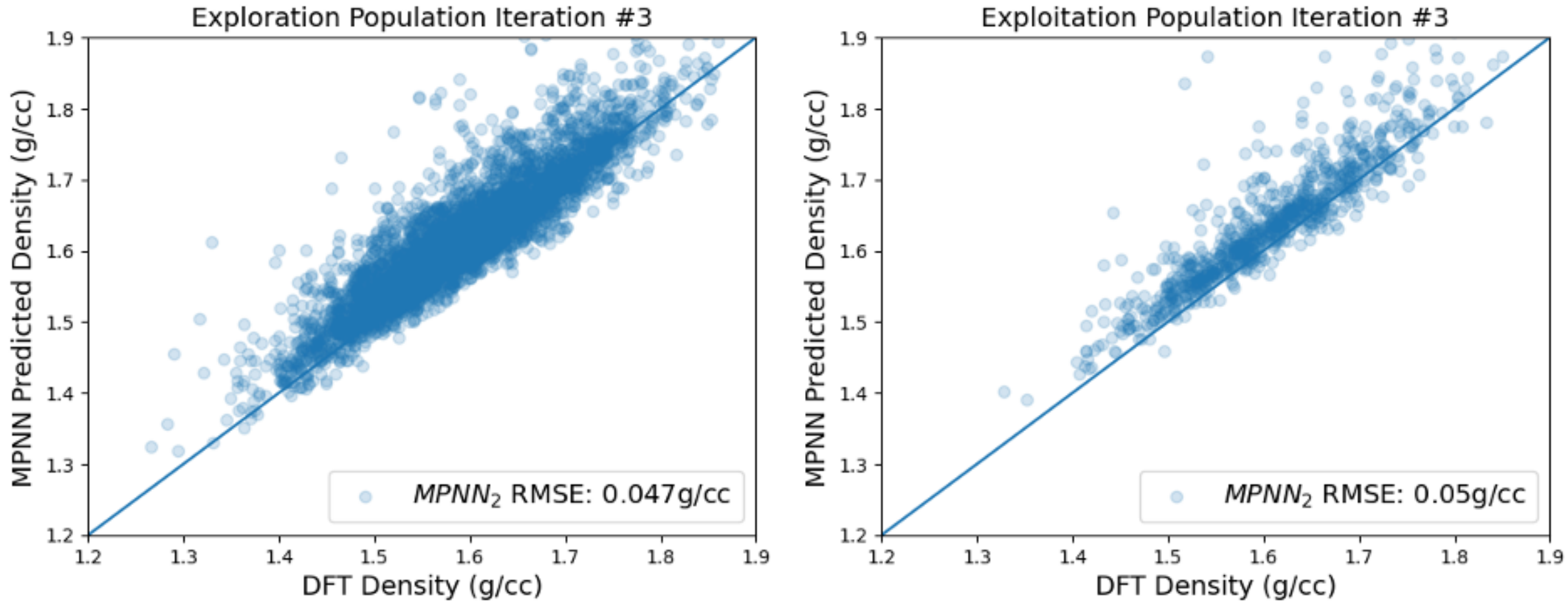

**Figure S16.** Density predictions of the $MPNN_2$ model (trained after two iterations of active learning) on the molecules arising from the exploration population (left) and exploitation population (right) from the JANUS generative model. Molecules are assigned to the population in which they first appear in the molecule generation process.

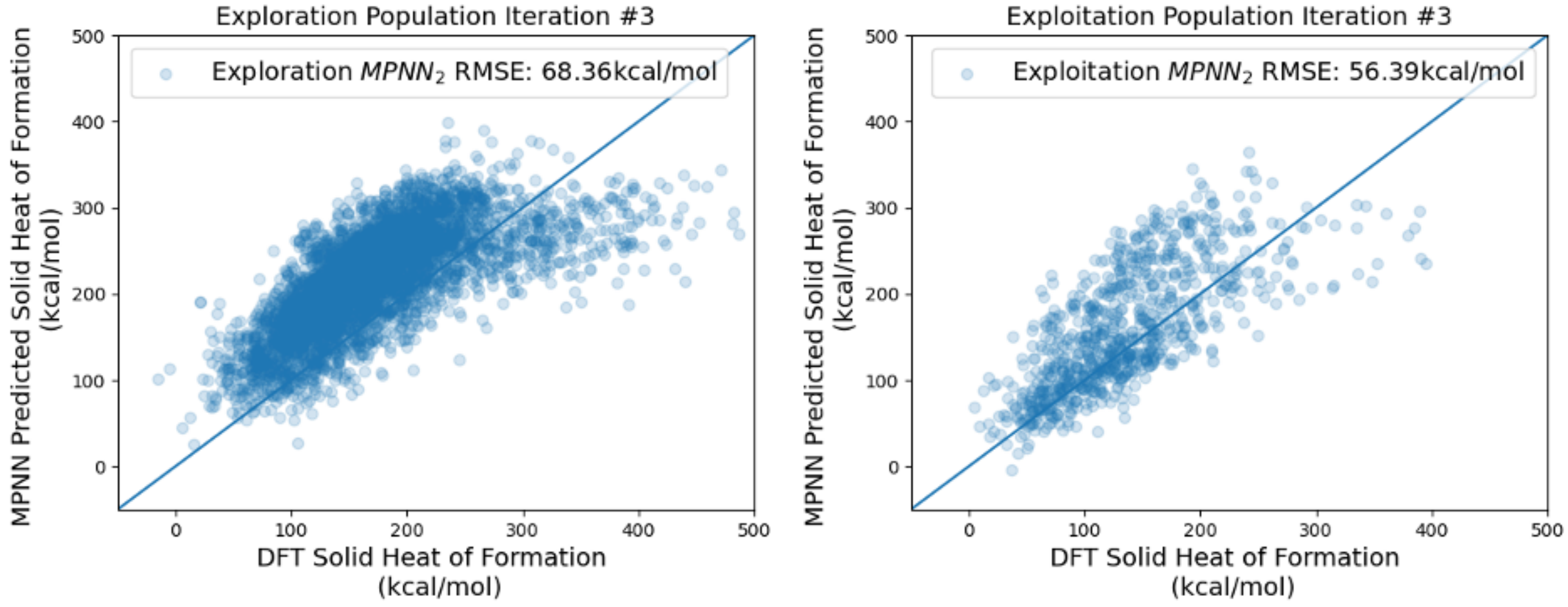

**Figure S17.** Solid heat of formation predictions of the $MPNN_2$ model (trained after two iterations of active learning) on the molecules arising from the exploration population (left) and exploitation population (right) from the JANUS generative model. Molecules are assigned to the population in which they first appear in the molecule generation process.

To further explore the out-of-distribution challenges of the property prediction models, we evaluate the MPNN models separately on the exploration and exploitation populations created by JANUS (Figure S16,S17). We perform this experiment on the third iteration of active learning since it has the most generated molecules. Although it might be expected that the MPNN model will perform much worse on the exploration population, we do not see a significant difference in the density predictions (Figure S16) and a slight increase in RMSE for the solid heat of formation predictions on the exploration population (Figure S17). Interestingly, there is a systematic underprediction of the values for the molecules with high solid heat of formation values in the exploration population- further underscoring the challenges with extrapolating to property values beyond the training distribution.

**Impact of Acquisition Function on Predictive Model Performance**

We also perform ablation experiments to understand how the choice of acquisition function impacts the performance of our generative models. Specifically, we explore how different means of sampling molecules from the first 3 active learning iterations impact the MPNN performance on a held out test set of generated molecules. For all sampling strategies, we sample N=[1000, 2500, and 5000] generated molecules from Iterations #1-3 to include in MPNN training. We explore the following three acquisition functions: i) **Random sampling:** molecules are sampled at random, ii) **Chemical Diversity:** molecules are sampled to maximize their chemical diversity. In this experiment, all molecules are represented with RDKit descriptors and sorted according to their average Tanimoto similarity to all other molecules in Iterations #1-3. The top N molecules with the highest average dissimilarity to all other molecules are sampled, iii) **Top Molecules:** Sampling N molecules with the highest DFT density and solid heat of formation values (akin to greedy sampling), as calculated by Equation 2. In all cases, we evaluate model performance against a held-out test set of 500 molecules that have been randomly sampled from Iterations #1-3.

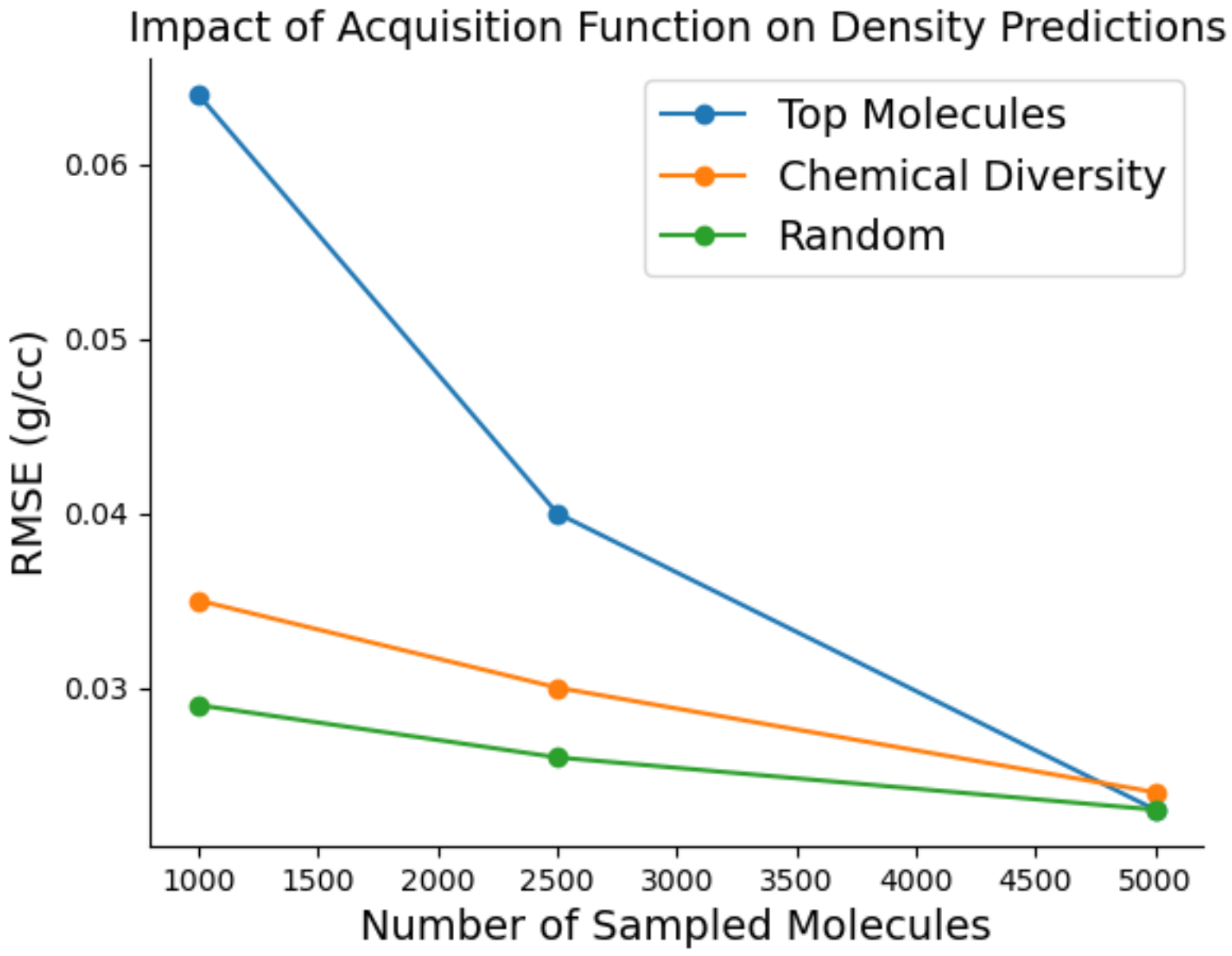

**Figure S18.** Impact of acquisition function on the accuracy of MPNN predictions of molecular density. MPNN models are evaluated on the same test set of 500 molecules that have been randomly sampled from Iterations #1-3. MPNN models are trained on N=[1000, 2500 or 5000] generated molecules from Iterations #1-3 that are either sampled according to either choosing the molecules with the top properties (blue), chemical diversity (orange), or at random (green), as detailed above.

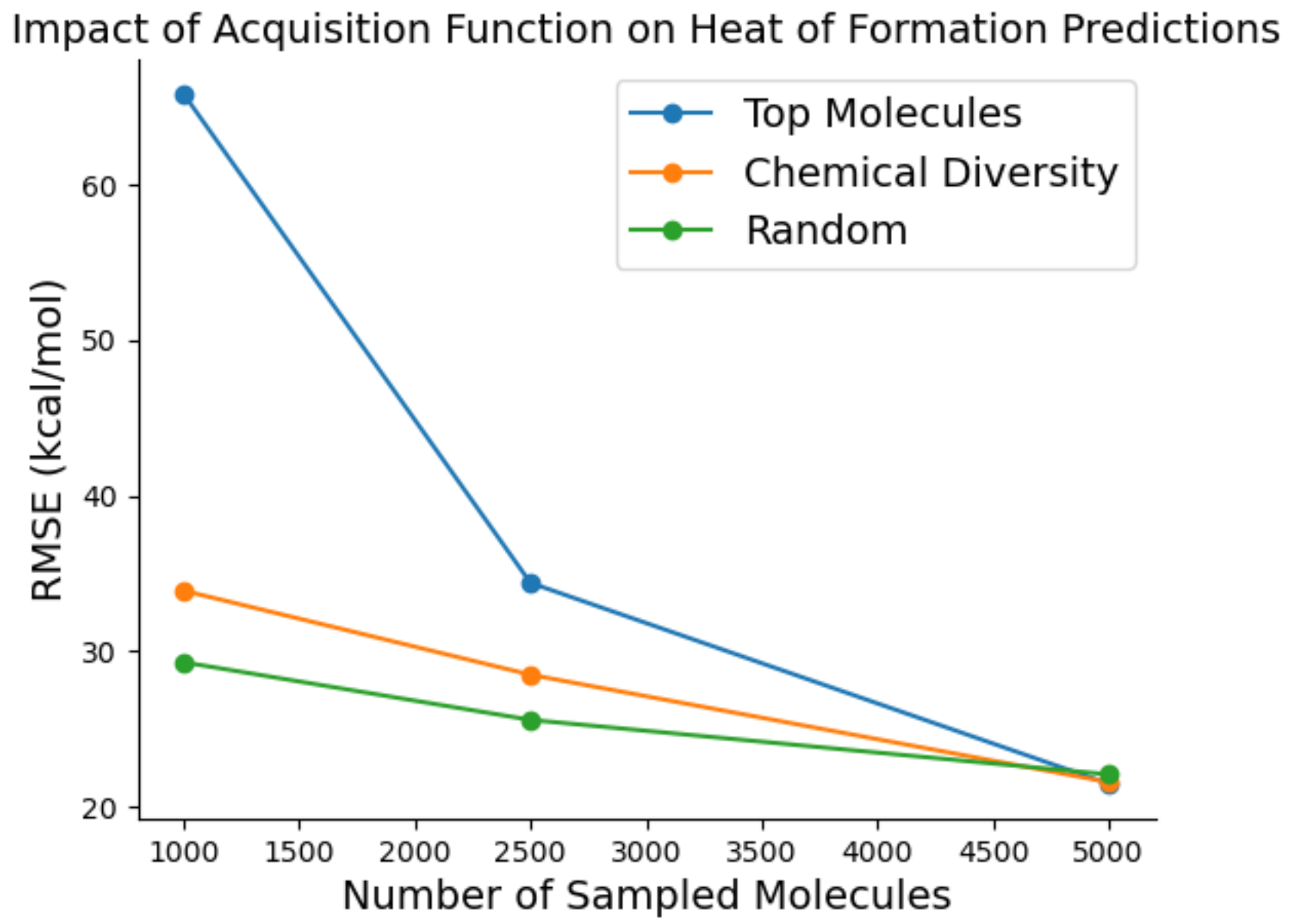

**Figure S19.** Impact of acquisition function on the accuracy of MPNN predictions of solid heat of formation. MPNN models are evaluated on the same test set of 500 molecules that have been randomly sampled from Iterations #1-3. MPNN models are trained on N=[1000, 2500 or 5000] generated molecules from Iterations #1-3 that are either sampled according to either choosing the molecules with the top properties (blue), chemical diversity (orange), or at random (green), as detailed above.

From these experiments, we find that randomly sampling molecules, as was done in the main text, leads to MPNN models with the best performance, whereas only sampling the top molecules achieves the worst performance. The poor performance achieved by only sampling top molecules is unsurprising since the resulting MPNN models would likely overfit on these top molecules and fail to be trained on a diversity of molecular structures. Although random sampling and diversity-based sampling give comparable performance, the slightly worse performance of diversity-based sampling may be due to its undersampling of well-populated regions of chemical space. All three sampling methods give nearly identical performance after N=5000 molecules have been sampled, indicating that chemical space has been thoroughly covered.